\documentclass[11pt,a4paper]{article}
\usepackage[hyperref]{main}
\usepackage{times}
\usepackage{latexsym}
\usepackage[utf8]{inputenc} 
\usepackage[T1]{fontenc}    
\usepackage{hyperref}       
\usepackage{url}            
\usepackage{booktabs}       
\usepackage{amsfonts}       
\usepackage{nicefrac}       
\usepackage{microtype}      
\usepackage{soul}
\usepackage{wrapfig}
\usepackage{multirow}

\usepackage{amsmath,amsfonts,bm}

\def\eqref#1{equation~\ref{#1}}

\def\1{\bm{1}}

\DeclareMathAlphabet{\mathsfit}{\encodingdefault}{\sfdefault}{m}{sl}
\SetMathAlphabet{\mathsfit}{bold}{\encodingdefault}{\sfdefault}{bx}{n}

\DeclareMathOperator*{\argmax}{arg\,max}

\usepackage{graphicx} 
\usepackage{bm}       
\usepackage{xcolor}   
\usepackage[nameinlink]{cleveref} 

\usepackage[normalem]{ulem} 
\usepackage{subcaption} 

\aclfinaltrue

\title{Self-play for Data Efficient Language Acquisition}

\author{%
  Charles Lovering \\
  Brown University\\
  \texttt{charles\_lovering@brown.edu} \\
  \And
  Ellie Pavlick \\
  Brown University\\
  \texttt{ellie\_pavlick@brown.edu} \\
}

\begin{document}

\maketitle

\begin{abstract}
When communicating, people behave consistently across conversational roles: People understand the words they say and are able to produce the words they hear. To date, artificial agents developed for language tasks have lacked such symmetry, meaning agents trained to produce language are unable to understand it and vice-versa. In this work, we exploit the symmetric nature of communication in order to improve both the efficiency and quality of language acquisition in learning agents. Specifically, we consider the setting in which an agent must learn to both understand and generate words in an existing language, but with the assumption that access to interaction with ``oracle'' speakers of the language is very limited. We show that using self-play as a substitute for direct supervision enables the agent to transfer its knowledge across roles (e.g. training as a listener but testing as a speaker) and make better inferences about the ground truth lexicon using only a handful of interactions with the oracle. 
\end{abstract}

\section{Introduction}
{As artificially intelligent agents become more pervasive in society, our ability to communicate with them becomes increasingly important. Many of the applications in which we'd like to interact with AI agents, e.g.\ in-home assistive robotics or search and rescue teams, benefit when agents can both understand and generate natural language, e.g. to follow instructions \cite{anderson2018vision} as well as to communicate their own state or uncertainty \cite{knepper2015recovering,whitney17}. Currently, however, training agents to communicate in this way is a data-intensive process, which would likely need to be repeated each time new objects are introduced into the environment or new words into the vocabulary. Moreover, most existing work assumes that agents always adopt the same conversational role, i.e. they are trained only to speak or only to listen, or are trained to speak using a vocabulary disjoint from the vocabulary they can understand \cite{kottur-etal-2017-natural, das2017learning}.} 

{In this work, we seek to train an agent from scratch to both generate and comprehend a new language, but assuming access to a speaker of the language is very limited, both in terms of the number of training examples and the type of supervision. This setting matches the constraints we'd expect if we were to ask non-experts to train systems in realtime: Most people would be willing to describe a few objects in their house (but not every single one, and not 100s of times each), and would likely be more willing to train via speaking (requesting a coffee) than via listening (fetching a requested coffee). We use a simulated setting to investigate the use of self-play as a substitute for direct interaction with speakers of the language, and show that agents trained this way can learn a lexicon that generalizes to novel items and novel conversational roles.}

{\textbf{Hypothesis.} Self-play in a reference game setting enables agents to transfer learned information across conversational roles.}

\textbf{Contributions.} While prior work has investigated the use of self-play as a substitute for supervision in general \cite{silver2017mastering} and in language games specifically \cite{lowe2020interaction}, this is the first work to our knowledge that investigates whether it is possible to for agents to perform both conversational roles (speaking and listening) despite only receiving supervision in one role. We analyze performance both quantitatively (task accuracy) and qualitatively (visualizations of lexicons), and draw three main conclusions. (1) Agents trained with self-play but without direct supervision learn a separate lexicon for speaking than for listening (\S\ref{sec:emergent}). (2) Using self-play, agents can learn to speak despite only receiving direct supervision as a listener (and vice-versa), though the ability to transfer across roles may be diminished when we account for effects such as natural language pragmatics (\S\ref{sec:oracle}). (3) When direct supervision is further limited (e.g.\ 10s of interactions total), self-play can enable agents to reach near-perfect accuracy, but requires stronger assumptions about the quality of the training interactions (\S\ref{sec:limited}).\footnote{Our code is available at \url{http://bit.ly/self-play}. See Appendix A for more information.}

\section{Related Work}
Most closely related to our work is Lowe et. al \cite{lowe2020interaction}, which provides an in-depth analysis of self-play and supervision, but does not address the question of generalization across speaker-listener roles. Overall, our conclusions are consistent with theirs, although we find that the agents are able to perform well without adopting various learning schedules (which was a focus of their work). Furthermore, our results in Section \ref{sec:limited} highlight that the teacher loss they employ is crucial for performance in limited data settings. Also relevant is \cite{gupta2019seeded}, which proposed the use of self-play among a population of agents as a means of cultivating agents with language learning priors, echoing a line of work \cite{livska2018memorize, frankle2018lottery} showing that neural networks' efficacy is in large part due to initialization, not only training.

Generally speaking, work in \textit{emergent communication} \cite{das2017learning, lazaridou2018emergence} analyzes agents that develop a shared protocol by playing reference games  \cite{lewis2008convention}. Kottur et. al \cite{kottur-etal-2017-natural} present results showing that computational models do not learn compositional protocols by default. Instead, the agents tend to develop brittle protocols that have trouble generalizing to novel items. Several approaches have been proposed which could encourage models to learn more general representations of language, including compression \cite{kirby2015compression}, efficiency \cite{gibson2019efficiency}, memory constraints \cite{kottur-etal-2017-natural}, pragmatic constraints \cite{tomlin2018incremental}, and positive biases to respond to other agents \cite{jaques2018intrinsic, eccles2019biases}. Some work, like ours, assumes access to symbolic representations of referents and their attributes, whereas others' are set in pixel-based multi-agent games \cite{jaques2018intrinsic, eccles2019biases, das2018tarmac} or multi-agent grid-worlds \cite{sukhbaatar2016learning, lowe2019pitfalls}.
Our work also relates to a broader body of work on speaker-listener models, specifically pragmatically-informed models in which speakers reason recursively about listeners' beliefs (and vice-versa) \cite{frank2012predicting,goodman2016pragmatic}. Such models have been used in applications such image captioning \cite{andreas-klein:2016:EMNLP2016,yu2017joint,monroe2015learning}, and robotics \cite{vogel2010learning,vogel-EtAl:2013:NAACL-HLT,fried2018speaker}. Lastly, the experimental setup in this work is quite similar to that off applied settings in dialog systems \cite{shah2018bootstrapping, ghandeharioun2019approximating}.
\section{Framework}
\subsection{Game} 
Following prior work \cite{kottur-etal-2017-natural, lazaridou2018emergence, lowe2020interaction}, we use a reference game setting \cite{lewis2008convention} where a speaker observes an item (the referent) and describes it to a listener, who tries to select the referent from among a set of distractors. Like \cite{lowe2020interaction}, we formalize this game as a Partially Observable Markov Decision Process \cite{kaelbling1998planning}: given a state $s \in S$, an agent (bot) $b$ receives an observation $o$; $o$ depends on both the state and $b$'s role in the game (speaker or listener). Then the agent \textit{b} performs actions $a \in A$, and finally, receives a reward $r \in \{-1, 1\}$. The state comprises the referent and the distractors. For the listener agent, the state comprises the speaker's message as well. A speaker $b_s$ observes the referent $t$ and produces a message (its sequence of actions) $M \in A^w$, where $w$ is the length of the message produced in words. A listener $b_l$ observes the referent and distractors (also called the context) in a random order and the speaker's message $M$. The listener then selects an item $i$ in context and both agents are rewarded $r = \1(t = i) - \1(t \ne i)$. In our experiments agents are trained in a \textit{target} role (speaker or listener) and then tested in the alternate \textit{novel} role.

We use two datasets, intended to capture qualitatively different properties of language and reference. First, \textit{Shapes}, similar to datasets used in previous work \cite{kottur-etal-2017-natural, lazaridou2018emergence}, consists of items with three attributes (e.g. \textit{color}, \textit{style}, \textit{shape}), where each attribute can take one of ten values (e.g. \textit{red}, \textit{dark}, \textit{triangle}). {All attribute values are uniformly distributed. Thus, \textit{Shapes} exemplifies simple, compositional structure in which a language that assigns unique words to each attribute-value pair will perform perfectly.} Second, like in previous work \cite{lazaridou2018emergence}, we use Visual Attributes for Concepts (\textit{Concepts}) \cite{silberer2013models}. \textit{Concepts} consists of items, each with $597$ binary attributes that correspond to real-world concepts. For example, \textit{pelican} includes attributes like \textit{has long neck, flies}, whereas \textit{fork} has attributes like \textit{made of steel, has prongs}. Thus, there is a co-occurence structure to the attributes that the agents can use to learn higher-level lexical concepts (\textit{animal, tool}).

For \textit{Shapes}, {our train/dev/test splits are 800/100/100, respectively.} We set the vocabulary size $|V| = 30$, allowing for a one-to-one correspondence between words and attributes {(3 attributes $\times$ 10 values per attribute)}. For  \textit{Concepts}, we {use splits of 455/55/55, and} set the vocabulary size $|V| = 100$. For both datasets, all test items are unobserved combinations of attributes.

\subsection{Learning Agents}
\label{sec:agents}
\textbf{Architecture.}
Agents consist of four modules: item $E_{item}: \mathcal{D} \to R^{\text{d}}$ and word  $E_{word}: V \to R^{\text{d}}$ embeddings, a message decoder $S: R^{\text{d}} \to V \times \dots \times V$, and a message encoder $L: R^{\text{d}} \times \dots \times R^{\text{d}} \to R^{\text{d}}$ where $\mathcal{D}$ is the set of and $d = 64$ is the dimension of all embeddings and hidden states. We use linear projections for the embeddings, and experiment with LSTMs \cite{hochreiter1997long} and Transformers \cite{vaswani2017attention} for the encoders and decoders.

Agents speak by embedding the referent and then decoding the representation into a message. Agents listen by embedding each word in the speaker's message and then encoding these embeddings into a single representation. Next, the agent embeds each item in context, and matrix multiplies its message representation and the embedded context, producing a distribution over the items in context. During training, items are sampled categorically; during test, $\argmax$ is used.

\textbf{Training.} Agents are trained using a score function estimator \cite{williams1992simple} and optimized using RMSProp \cite{Tieleman2012}. An agent $b$ with $\pi_b$ and parameters $\theta_b$ (the parameters are all update-able parameters in its modules), for a given action $a$ and observation $o$, is assigned the following loss: $ -\log \pi_{b}(a \,|\, o, b_\theta) \times (r - \hat{l}). $ Thus, each action (producing a word or selecting an item) is either rewarded or punished by $r$. The running mean of previous losses $\hat{l}$ is used as a baseline to reduce variance.

\subsection{Oracle Agents}
We are interested in the ability of learning agents (\S\ref{sec:agents}) to acquire an existing language (eventually, natural language). Thus, we define simulated ``oracle'' agents which deterministically produce messages from a ground-truth lexicon (mappings between words and attributes). This setup allows us to inspect how well the learning agents recover the desired lexicon.

For \textit{Shapes}, the oracles use a one-to-one mapping between attributes and words. For ease of interpretation, we set to the mapping to be the diagonal (such that the $k$th word corresponds to the $k$th attribute). The oracles produce words for attributes in a deterministic order, but they ignore the order of incoming messages, choosing the item in context whose features most overlap with the message.

For \textit{Concepts}, the oracle agents are rational interlocutors which incrementally build up messages. (Specifically, our oracle speaker behaves like the S1 ``rational speaker'' as defined in the Rational Speech Acts (RSA) model \cite{goodman2016pragmatic}; see Appendix B for details.) The listener oracle uses the same mechanism as the listener oracle for \textit{Shapes}: It chooses the item for which entries in the lexicon most overlap with the message received. Furthermore, we let the oracle agents use category words (e.g. \textit{animal}) in addition to attribute words (e.g. \textit{has fur}) when producing messages, providing an additional challenge (and opportunity) to the learning agents: The learning agents must ideally learn to associate sets of features with the item categories, while also reasoning about generic attributes, which are true of many items and thus likely to be under-produced by the oracle.
\subsection{Evaluation Metrics and Visualizations} 
We use both quantitative and qualitative metrics to determine whether a learning agent has acquired the target language. Quantitatively, we look at overall accuracy on the reference game task. Accuracies are always computed over heldout test sets in which all items are combinations of attributes unobserved during training. Qualitatively, we visualize agent's lexicons, i.e. which words they produce for a given attribute, by generating heatmaps. 

\textbf{How to interpret lexicons/heatmaps:} In our lexicons (Figs.\ \ref{fig:emergent}, \ref{fig:oracle-lex}, \ref{fig:oracle-limited-lexicons}), the columns are the attributes of the referents, and the rows are the words that speakers can produce. Each cell $ij$ shows, given that the referent had attribute $j$, the number of times the agent produced the word $i$. Because each message has multiple words, some of the co-occurences are due to the other attributes in the referent item, leading to some patterns of ambient association. We cannot inspect heatmaps over \textit{Concepts} as the figures are very large (597 attributes); instead we examine example dialogs.

\subsection{Self-play} 
\textbf{Motivation.} Ideally, we can train agents to speak, to listen, (or both) as needed. However, if training comes from interaction with a human, data may be limited by role and by amount. Thus, we ask whether self-play can leverage that limited data. This is why we do not directly train agents in the roles they are tested in. The target role performance serves as an upper-bound ``skyline'' benchmark. (The self-playing agents sometimes match or exceed this ``skyline'' performance, but its not truly a valid comparison because our problem setting assumes direct training is not an option.).

\begin{figure}
\centering
\includegraphics[width=0.45\textwidth]{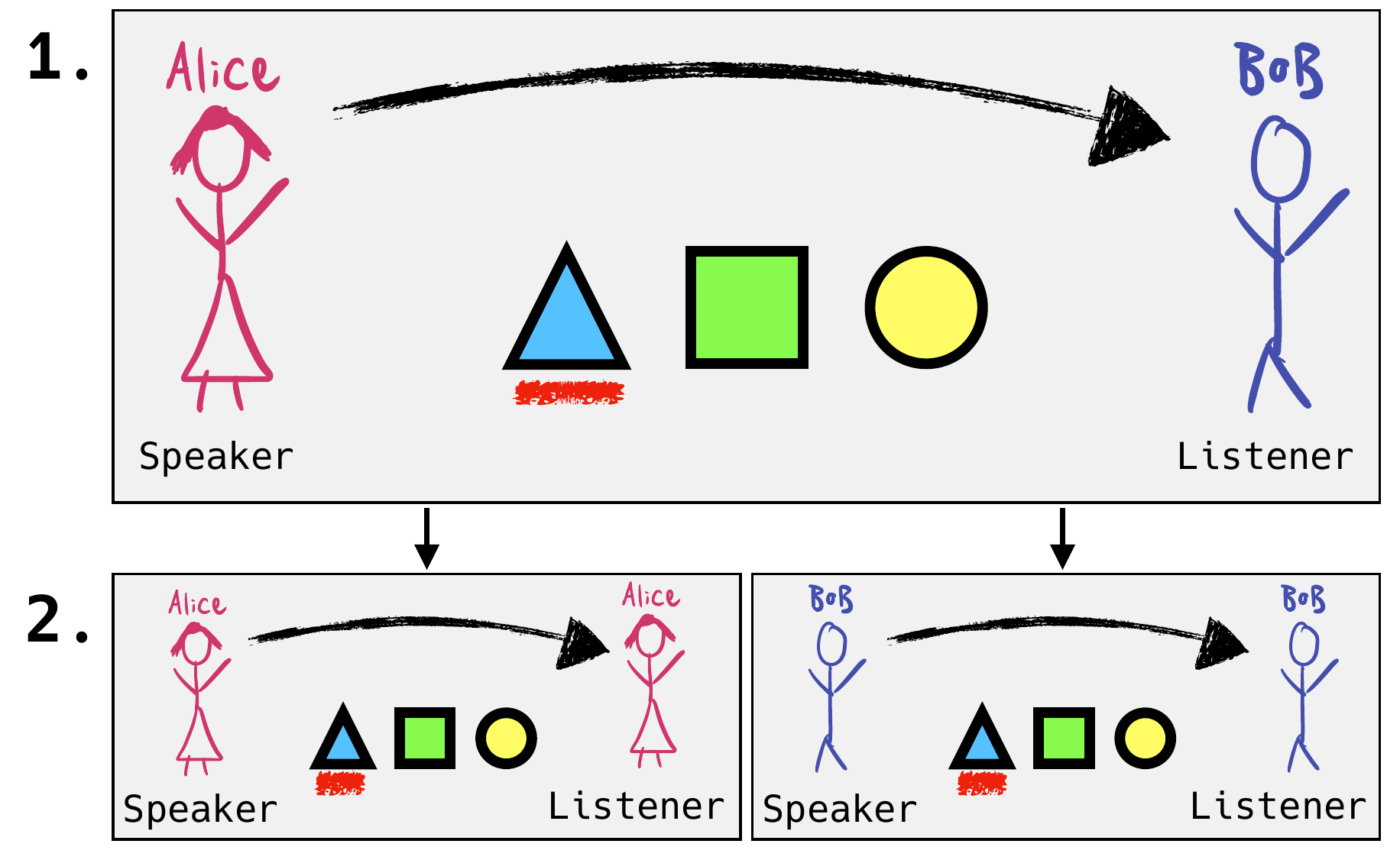}
\caption{\textbf{Reference game followed by self-play.}  Alice wants the blue triangle, and asks Bob to get it for her. After the agents (1) play the game in their target roles, they (2) self-play and practice in both roles  \textit{in isolation} using the same objective function and data.}
\label{figure:ref-game}
\end{figure} 

\textbf{Summary}. Fig.\ \ref{figure:ref-game} summarizes how we use self-play. Two agents with disjoint parameters, each have speaking and listening modules. We call interaction between the agents ``direct interaction'' and interaction between an agent's own modules ``self-play''. Direct interaction is limited between agents to their respective target roles. The idea is that in some scenarios interaction in the novel role is unavailable (say with a human only interested in talking), and we would like an agent to learn from this one-sided interaction (e.g.\ only listening). To that end, after each interaction between the agents in their target roles, agents self-play. 

\textbf{Self-play technical definition.} A speaker $b$ sends a message $M_{\text{direct}}$ to a listener and is rewarded $r_{\text{direct}}$ if the listener selects the referent. Then $b$ self-plays, taking both roles and receives a reward $r_{\text{self}}$ in both roles. Each action (the choice of a word or an item) is rewarded $r_{*}$ using a score-function estimator. The running mean of previous losses $\hat{l}$ is used as a baseline to reduce variance. This is the overall loss for the speaker agent:
\begin{align*}
    &\underbrace{\sum_{a \in M_{\text{direct}}} -\log \pi^S_{b}(a \,|\, o, \theta_{b})  (r_{\text{direct}} - \hat{l})}_{\text{Speaker loss, direct.}}\\
 &+ \underbrace{\sum_{a \in M_{\text{self}}} -\log \pi^S_{b}(a \,|\, o, \theta_{b})  (r_{\text{self}} - \hat{l})}_{\text{Speaker loss, self-play.}}\\
  &+  \underbrace{-\log \pi^L_{b}(a \,|\, o, \theta_{b})  (r_{\text{self}} - \hat{l})}_{\text{Listener loss, self-play.}} \label{eq:1}
\end{align*}

\section{Experiments}
\subsection{No Direct Supervision}
\label{sec:emergent}

We first consider the effects of self-play on cross-role generalization in the setting in which we have no access to direct supervision from an oracle. That is, in this setting, two learning agents are trained to communicate with one another in order to achieve high accuracy on the described reference game. This setting is known as \textit{emergent communication}, and has been widely studied  \cite[inter alia]{das2017learning,kottur-etal-2017-natural,cao2018emergent}, often with some hope that the learned protocols can be later ``translated'' to natural language for downstream applications. Recent work \cite{lowe2019pitfalls} has shown that emergent communication does \textit{not} provide a good initialization for later supervised learning. Here, we further show that emergent protocols do not result in grounded lexicons that generalize across speaker-listener roles.
\begin{table*}[ht!]
    \centering
  \begin{tabular}{@{}l|ccc|ccc|c@{}}
    \toprule
  	& \multicolumn{3}{|c}{Target Role} & \multicolumn{3}{|c|}{Novel Role} & $\Delta_{role}$ \\
   \midrule
         & $\%_{base}$ & $\%_{sp}$ & $\Delta_{sp}$  & $\%_{untrained}$ & $\%_{sp}$ & $\Delta_{sp}$ &   \\ 
   \midrule
\textit{Shapes}, LSTM & 93.4 & 96.7 & +3.3 & 21.1 & 96.3 & +75.2 & -0.4\\
\textit{Shapes}, Trans. & 97.3 & 99.4 & +2.1 & 20.1 & 98.8 & +78.7 & -1.4\\
\midrule
\textit{Concepts}, LSTM & 75.9 & 78.2 & +2.3 & 19.3 & 34.5 & +15.2 & -43.7\\ 
\textit{Concepts}, Trans. & 76.5 & 86.8 & +12.3 & 25.6 & 86.2 & +60.6 & -0.6\\
\bottomrule
\end{tabular}
\caption{\textbf{Test accuracies for agents trained without any direct supervision.} Target Role means agents are tested in the same roles they were trained in ; Novel Role means they are tested in opposite roles (i.e. if agents were trained such that A always spoke and B listened, they are tested such that B speaks and A listens). $\%_{base}$ is performance without self-play, $\%_{sp}$ is with self-play, $\Delta_{sp}$ is the change in performance when adding self-play, $\Delta_{role}$ is the drop in performance observed when moving to the novel role. For the Novel Role, the baseline approach is not competitive as it is (basically) untrained, so we denote this by $\%_{untrained}$.}
        \label{table:emergent}
\end{table*}

\begin{figure*}[ht!]
    \centering
    \begin{subfigure}[t]{0.33\textwidth}
        \centering
        \includegraphics[width=0.99\textwidth]{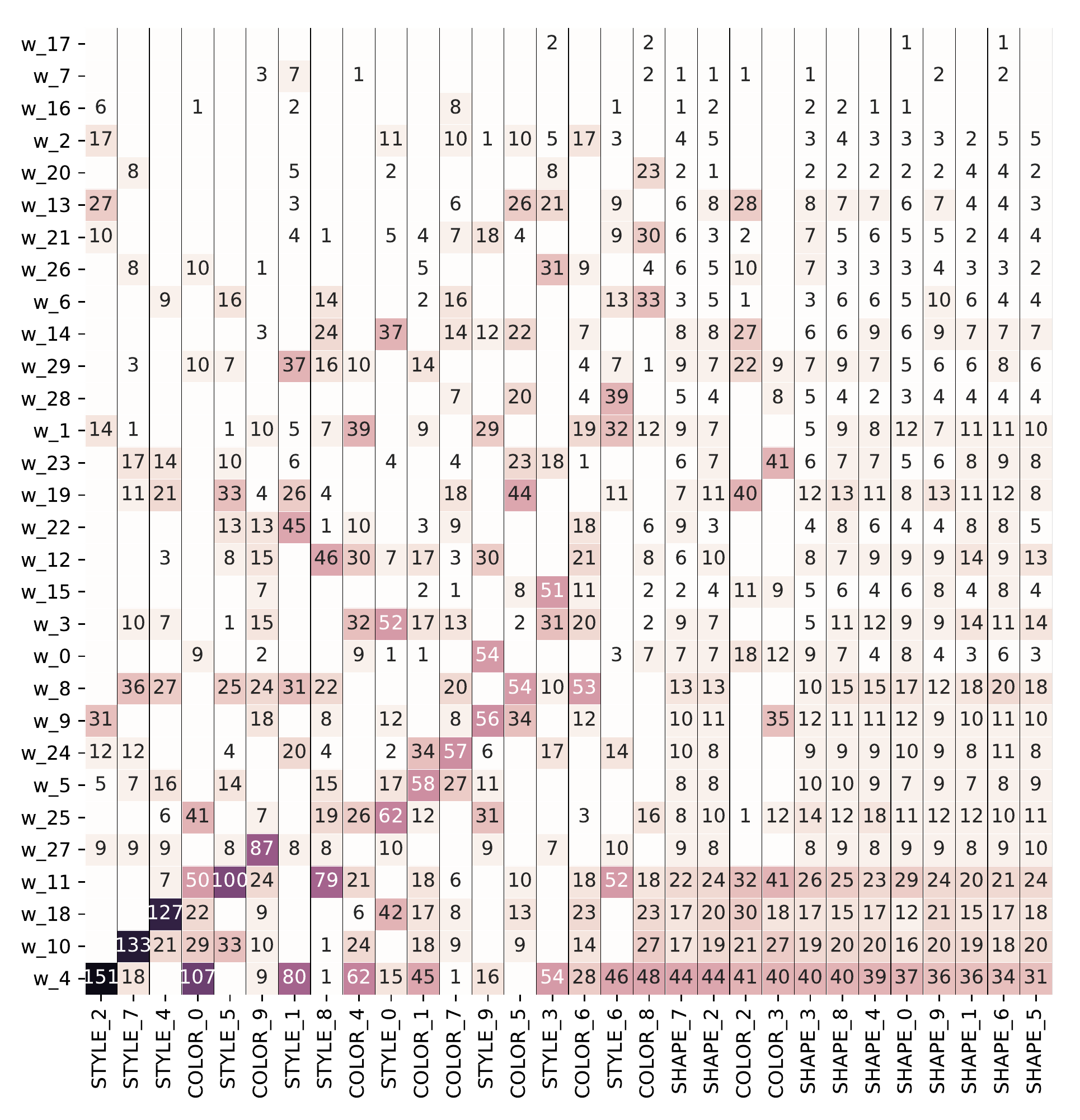}
        \caption{Target role lexicon.}
       \label{fig:emergent:target-lex}
    \end{subfigure}%
    ~ 
    \begin{subfigure}[t]{0.33\textwidth}
        \centering
        \includegraphics[width=0.99\textwidth]{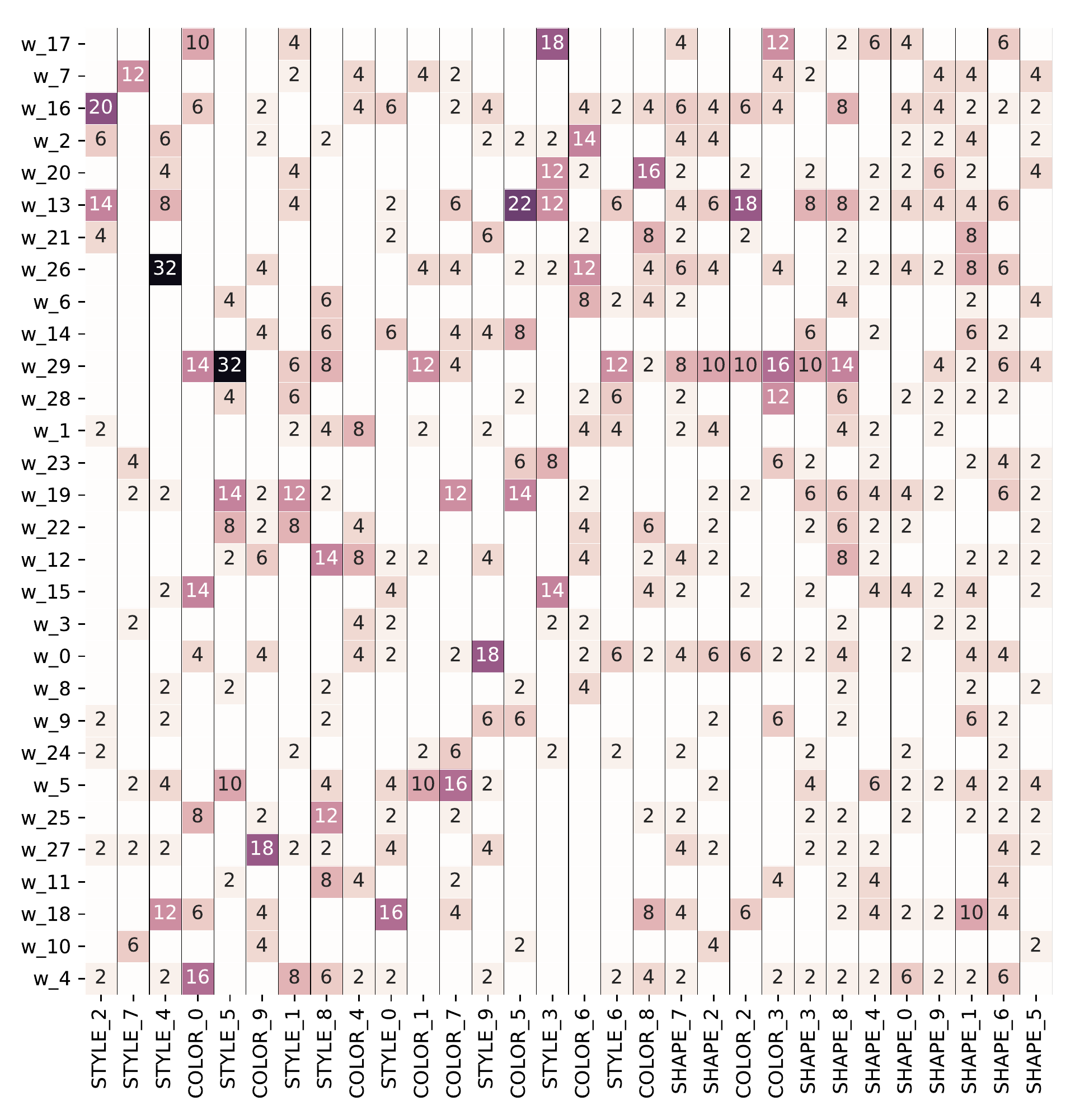}
        \caption{Novel role lexicon. }
         \label{fig:emergent:novel-lex}
    \end{subfigure}%
        ~ 
        \begin{subfigure}[t]{0.33\textwidth}
        \centering
     \includegraphics[width=0.99\textwidth]{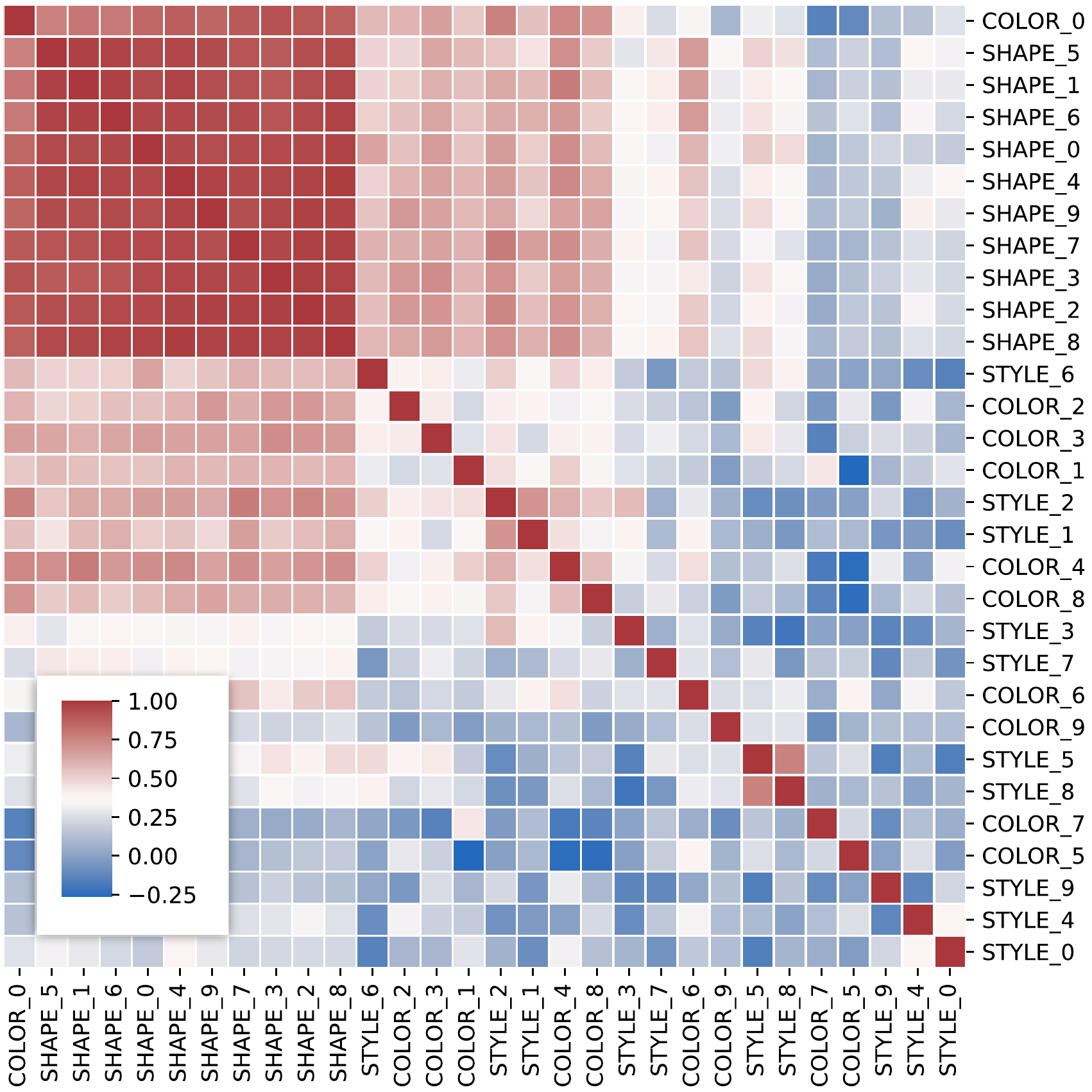}
        \caption{Target role message correlations.} 
         \label{fig:emergent:correlations}
    \end{subfigure}%
    ~
    \caption{\textbf{Emergent communication.} Visualizations of learned lexicons--i.e. mapping between words ($y$-axis) and attributes ($x$-axis). \textbf{(a)}: Lexicon used when agents are in their target roles (A speaking; B listening). \textbf{(b)}: Lexicon used when agents are in novel roles (A listening; B speaking). We see here that the distribution over attributes referred to by a given word differs depending on whether agents are in their target vs.\ novel roles. \textbf{(c)}: Correlation plot between messages produced for each attribute, for agents in their target roles. Cell $[i][j]$ shows the average overlap between messages produced for objects with attribute $i$ and those produced for objects with attribute $j$. We see that models learn to differentiate color and style attributes, but not shape attributes. }
    \label{fig:emergent}
\end{figure*}

\textbf{Agents achieve high cross-role performance, but do not learn cross-role lexicons.} Table \ref{table:emergent} shows the performance of agents in their target roles and in their novel roles. We can see that, with self-play, agents are able to communicate effectively even when tested in the roles opposite those in which they were trained. However, when we look at the lexicons the agents are using (Figs.\ \ref{fig:emergent:target-lex} and \ref{fig:emergent:novel-lex}), we can see that the {grounded} meanings of words are not constant across roles, i.e. the word \textit{``blue''} {maps to a different distribution over referents when used by agent A than when used by agent B}. Moreover, we see the previously-observed phenomenon \cite{kottur-etal-2017-natural} in which agents do not learn sparse lexicons (with one-to-one mappings between words and attributes) as we expect to see in natural languages. In particular, our agents converge to a protocol in which they distinguish between items by referring to color and style attributes, but never learn to differentiate between different shapes (Fig.\ \ref{fig:emergent:correlations}).

\subsection{Oracle Supervision}
\label{sec:oracle}
In practice, we are interested in teaching agents to use natural language, so that they can communicate with humans rather than just with one another. In our oracle supervision setting, we assume the learning agent can have unlimited interaction with an oracle speaker of the language, but only in one direction. That is, the agent trains on the task of fetching objects specified by an oracle speaker, but never trains on the task of asking for objects and observing what is fetched. Then, at test time, we evaluate how well the agent does at asking for objects. (The process is flipped for agents trained as speakers but tested as listeners.)

\begin{figure*}[ht!]
    \centering
    \begin{subfigure}[t]{0.4\textwidth}
        \centering
        \includegraphics[width=0.99\textwidth]{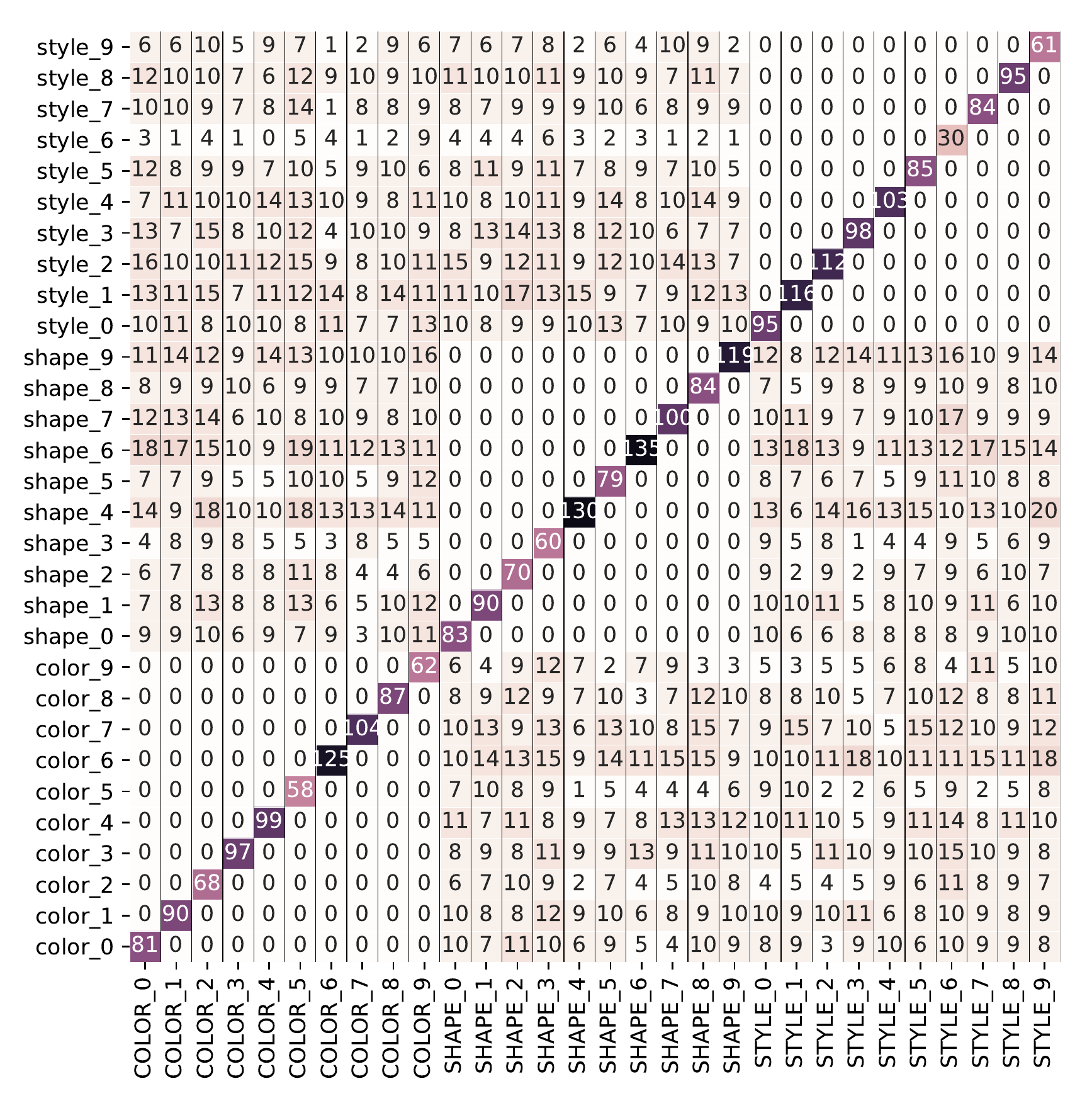}
        \caption{Oracle target task lexicon.}
    \end{subfigure}%
    ~ 
    \begin{subfigure}[t]{0.4\textwidth}
        \centering
        \includegraphics[width=0.99\textwidth]{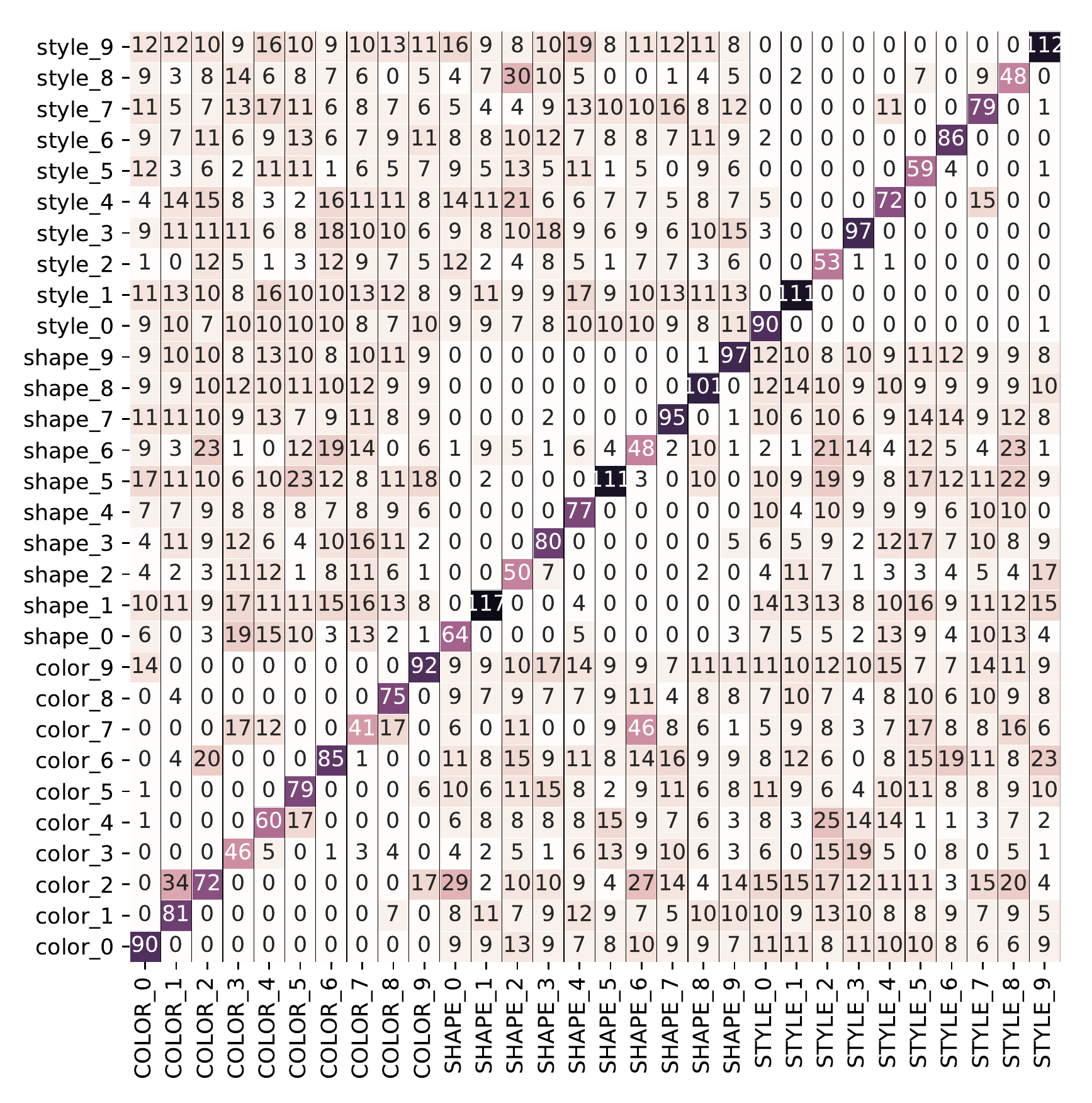}
        \caption{Oracle novel task lexicon. }
    \end{subfigure}%
    \caption{\textbf{Oracle supervision.} Visualizations of learned lexicons--i.e. mapping between words ($y$-axis) and attributes ($x$-axis). The diagonal shows that the agents recover the oracle lexicon. The grid blocks show that the agents learn that the different values of a given attribute are mutually exclusive. The fact that (a) and (b) are similar shows that the information was transferred across roles.}
    \label{fig:oracle-lex}
\end{figure*}

\begin{table*}[ht!]
    \centering
\begin{tabular}{@{}l|l|ccc|ccc|c@{}}
    \toprule
  	& \multicolumn{1}{|c}{Trained As} & \multicolumn{3}{|c|}{Target Role} & \multicolumn{3}{|c|}{Novel Role} & $\Delta_{role}$ \\
   \midrule
       && $\%_{base}$ & $\%_{sp}$ & $\Delta_{sp}$  & $\%_{untrained}$ & $\%_{sp}$ & $\Delta_{sp}$ &   \\ 
   \midrule
\textit{Shapes}, LSTM & Speaker & 99.2 & 99.4 & +0.2 & 21.0 & 100.0 & +79.0 & +0.8\\
\textit{Shapes}, Trans. &  Speaker & 98.8 & 99.6 & +0.8 & 17.5 & 100.0 & +82.5 & +1.2\\
\textit{Shapes}, LSTM & Listener & 100.0 & 100.0 & +0.0 & 18.4 & 93.9 & +75.5 & -6.1\\
\textit{Shapes}, Trans. &  Listener &100.0 & 100.0 & +0.0 & 24.9 & 100.0 & +75.1 & +0.0\\
\midrule
\textit{Concepts}, LSTM &  Speaker & 86.6 & 80.0 & -6.6 & 17.1 & 67.6 & +50.5 & -12.4\\ 
\textit{Concepts}, Trans. & Speaker & 82.6 & 89.0 & +6.4 & 24.3 & 88.4 & +60.6 & -0.6\\
\textit{Concepts}, LSTM &  Listener & 95.4 & 87.0 & -8.4 & 21.5 & 42.6 & +21.1 & -44.4\\ 
\textit{Concepts}, Trans. & Listener & 97.7 & 86.6 & -14.1 & 23.5 & 48.7 & +25.2 & -37.9 \\
\bottomrule
\end{tabular}
    \caption{\textbf{Test accuracies for agents trained with oracles.} Target Role means agents are tested in the same roles on which they were trained; Novel Role means they are tested in opposite roles (e.g. if the agent was trained as a speaker then novel role means listener and vice-versa). $\%_{base}$ is performance without self-play, $\%_{sp}$ is with self-play, $\Delta_{sp}$ is the change in performance when adding self-play, $\Delta_{role}$ is the change in performance when switching to the novel role. For the Novel Role, the baseline approach is not competitive as it is (basically) untrained, so we denote this by $\%_{untrained}$.}
        \label{table:oracle}
\end{table*}
\textbf{Agents trained via self-play converge to the ground-truth lexicon and transfer the learned information across roles.} As was the case in the emergent (no-direct-supervision) setting discussed in Section \ref{sec:emergent}, we see that the use of self-play enables the learning agent to achieve high task accuracy when transferring to novel roles (Table \ref{table:oracle}). Unlike the emergent setting, however, in the oracle setting, we observe the desired behavior in which the converged model uses the ground-truth lexicon (e.g., for \textit{Shapes}, a one-to-one mapping between words and attributes) and it uses this same lexicon for both roles (Fig.\ \ref{fig:oracle-lex}). The benefit of self-play is clear for both \textit{Shapes} and \textit{Concepts}, but whereas novel-role performance on \textit{Shapes} is near ceiling, performance on \textit{Concepts} is lower. Specifically, accuracy is only around 50\% for a transformer trained as a listener but tested as a speaker. This contrast is worth highlighting, since our \textit{Concepts} data incorporates additional (realistic) assumptions about the type of language supervision a model would receive if interacting with a human. That is, for \textit{Concepts}, the oracle speaker does not exhaustively enumerate the properties of an object, but rather mentions only those attributes that are most discriminating, and the oracle listener assumes the learning speaker to behave similarly. In this setting, transferring to a novel role is more challenging. Table \ref{tab:oracle-dialogs} shows several example dialogs that provide intuition for why this setting is challenging. For example, the same words may pick out different referents in different contexts, and oracle listeners are often able to understand messages even when they contain some incorrect words. These properties can make it harder to reverse-engineer the oracle's lexicon and thus apply it when in the novel role. Improving performance in settings such as this would be a interesting direction for future research.

\begin{table*}[ht!]
    \centering
    \small
\begin{tabular}{@{}lp{.4\linewidth}p{.4\linewidth}l@{}}
\toprule
 & Context & Message & Pred. \\\midrule
S$\to$S &  \{thermometer, leopard, shack, accordion, {\bf drapes} \} & [clothing, different colors*] & drapes \\
S$\to$L & \{raspberry, muzzle, drapes, alligator, {\bf earmuffs}\} & [clothing*, different colors*, is black*, is blue*, made of different fabrics*, $\dots$]  &  drapes \\
\midrule
L$\to$L &  \{cello, trolley, leopard, stereo, {\bf cockroach}\} & [animals*, eats*, has wings*, is black*, is brown*, is red*, is slender*, is small*] & cockroach \\
L$\to$S & \{socks, {\bf cockroach}, crowbar, pine, orange\} & [eats grass, has feet, is red*, made of different fabrics, tools]  &  crowbar \\
\bottomrule
\end{tabular}
    \caption{\textbf{Example dialogs.} Produced by transformer agents trained using self-play with oracles supervision using the \textit{Concepts} data. Target in the context is bolded. S$\to$S denotes that a learning agent trained as a speaker is speaking (and the oracle is listening); S$\to$L denotes that a learning agent trained as a speaker is listening (oracle speaking). $*$ denotes that the word corresponds to a ground-truth attribute of the referent. Some messages shown have fewer than 10 words; this denotes that the message contained repeated words.} 
        \label{tab:oracle-dialogs}
\end{table*}
\subsection{Limited Oracle Supervision}
\label{sec:limited}
Lastly, we consider the case in which the agent's access to the oracle is limited not just in that it is restricted to a single role (speaking or listening) but also in terms of the total number of interactions. We envision a scenario like the following:
``Sam the student (the learning bot) works with Tetra the teacher (the oracle). Sam wants the red ball and asks \textit{``blue ball''}? Tetra gives Sam the blue ball, not the red ball, so Sam knows he said something wrong. Tetra is willing to repeat this process a few times (but not all day) for a few items (but not every single toy).''

We consider two variations on the above setting. The first is the restricted setting we have used thus far in which the only signal Sam receives is binary feedback on whether or not his choice/message was the correct one. That is, Sam does not receive feedback about what a better action (message or choice of object) would have been, and cannot remember the interactions so cannot recall anything about the oracle behavior during self-play. Since this setting is somewhat artificially constrained, we also introduce a second setting, \textbf{teacher loss}, in which Sam is able to have more substantive interactions with Tetra. Specifically, we assume that Tetra is able to give feedback beyond a simple yes/no--i.e. if Sam gives the wrong message, Tetra can say what the right one is. In addition, we assume Sam can remember Tetra's messages (``caching'' them) after interacting directly, and can use them later to improve self-play updates. Thus, we let agents in both roles train on these messages using a loss of negative log likelihood of producing the given message $M$: $\sum_{m \in M}  -\log \pi_s(m \,|\, t, \theta_b) \label{eq:2}$. Lowe et. al \cite{lowe2020interaction} use a similar loss. We emphasize that the agent only has access to a limited subset of the data with the oracle, and so has a comparatively small number of messages and contexts on which it can apply teacher loss. For the rest of the training data, the agent self-plays as normal.

We are interested in measuring how effective self-play is as a substitute for direct interaction with the oracle, as a function of the number of examples ($M$) and training iterations ($N$). Here, a single ``training example'' is a specific setting of the reference game (i.e. referent plus distractors), e.g. if $M=1$, the agent will only have seen a single message and a single set of items in which one was the correct one. The agent does not then have the opportunity to see this same message/item in the context of new distractors (as was the case in previous sections).  We vary $M \in \{5, 10, 50, 200, 800 \}$.  We consider two settings for $N$: the ``limited'' setting ($N=10$) in which the agent only gets to go over the $M$ examples a handful of times with the oracle, and the ``converged'' setting ($N=\infty$) in which the agent is able to repeatedly interact with the oracle on these $M$ examples as much as needed until converged. We hope to see only a small difference between the  ``limited'' and ``converged'' performances, indicating that the self-play was able to replace direct supervision.

\begin{figure*}[t!]
    \centering
    \begin{subfigure}[t]{0.5\textwidth}
        \centering
        \includegraphics[width=0.99\textwidth]{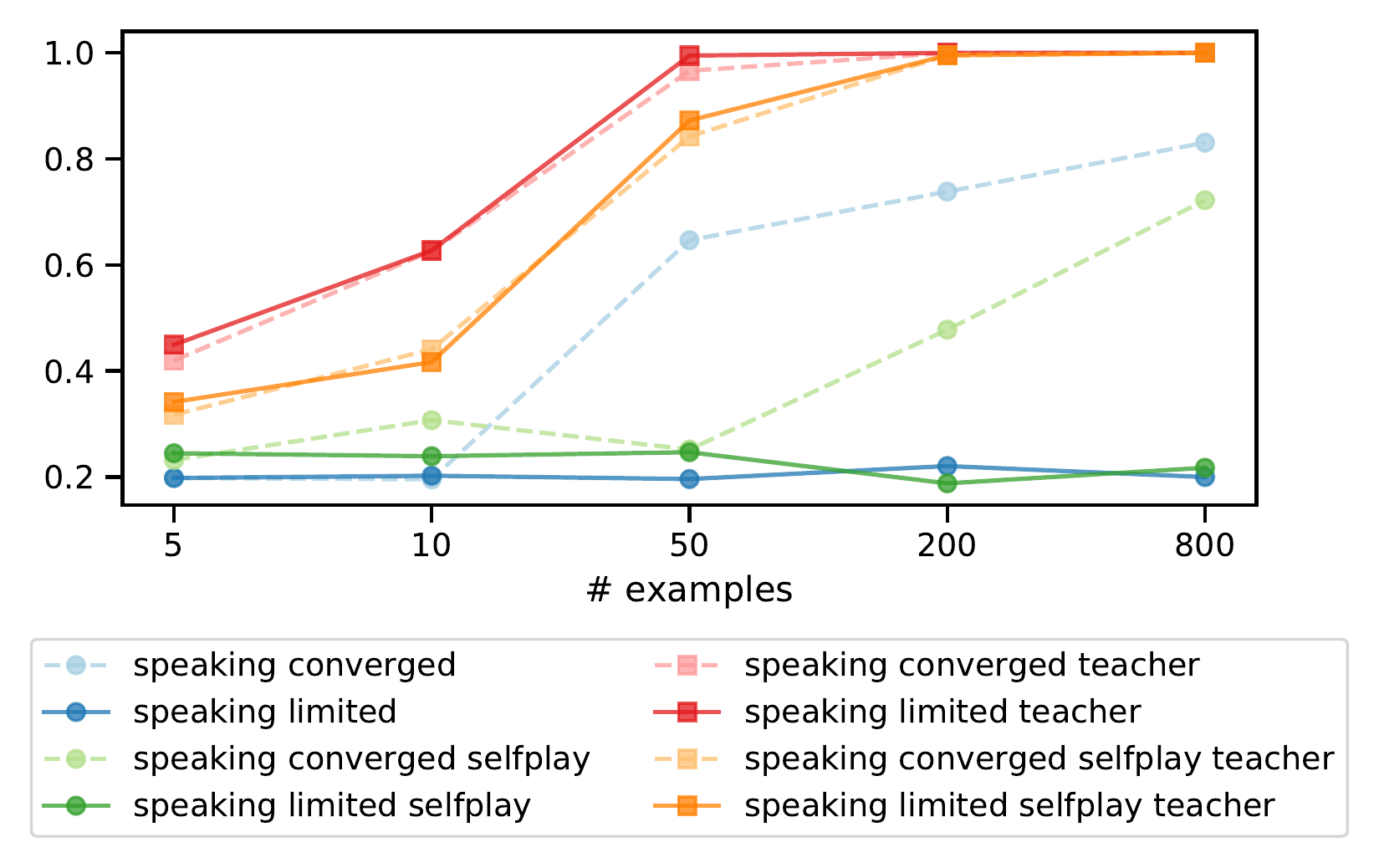}
        \caption{Target Role: Transformer, trained as a \underline{speaker}, \underline{speaking} with limited supervision across a range of settings.}
    \end{subfigure}%
    ~ 
    \begin{subfigure}[t]{0.5\textwidth}
        \centering
          \includegraphics[width=0.99\textwidth]{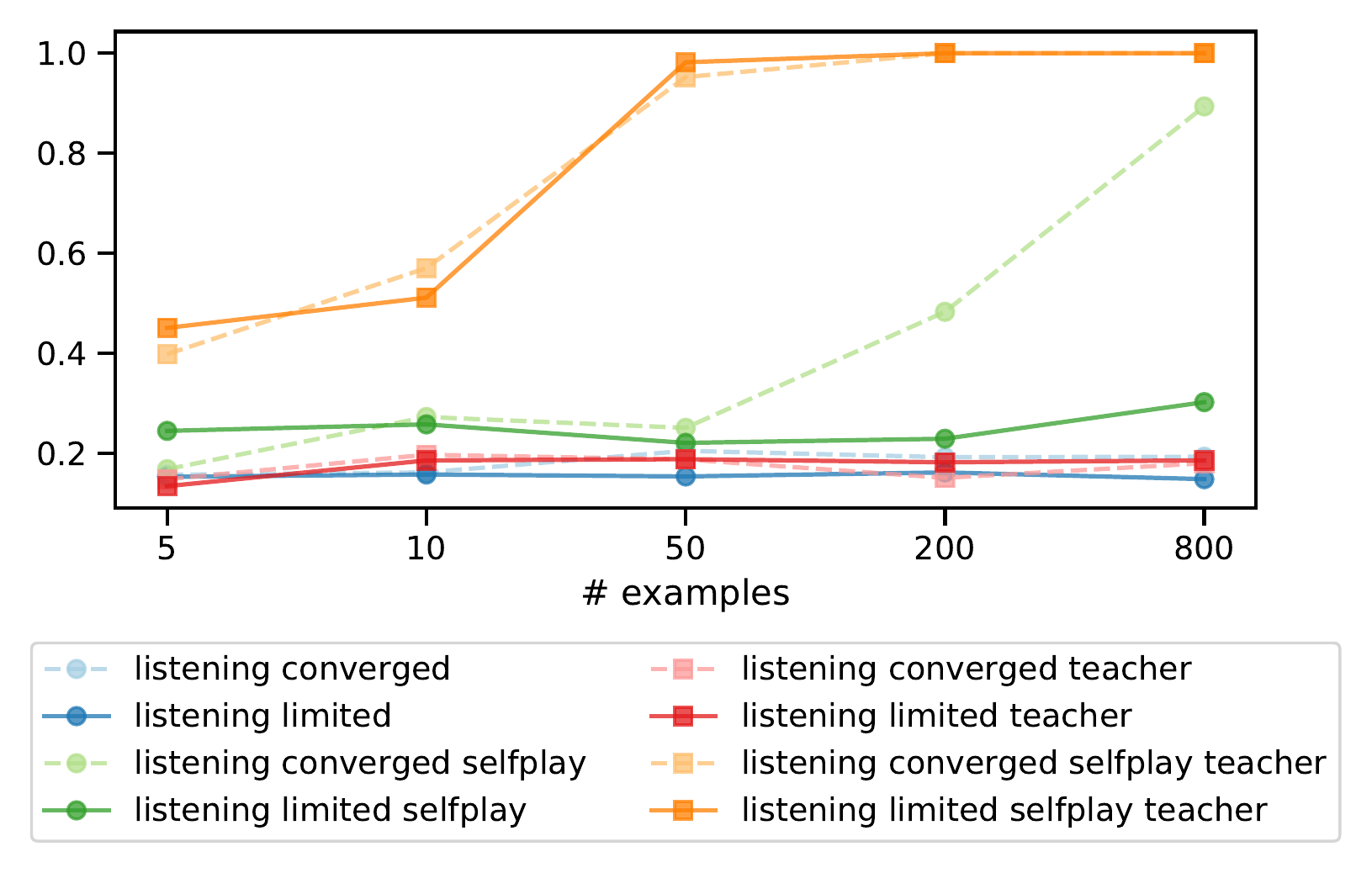}
        \caption{Novel Role: Transformer, trained as a \underline{speaker}, \underline{listening} with limited supervision across a range of settings.}
    \end{subfigure}
    \caption{\textbf{Teacher and self-play loss are critical for novel role performance.} {Comparison between models trained with/without teacher loss on target/novel roles.  Blue/green lines have no teacher loss; red lines have teacher loss but no self-play; orange lines have both teacher loss and self-play. Only the teacher+sp setting performs well when tested on the novel role.} (This is for \textit{Shapes} data, and where the target role is speaker, novel is listener; full results are in Appendix C.)}
    \label{fig:teacher-vs-no}
\end{figure*}
\textbf{In the limited data setting, self-play enables agents to transfer to novel roles, but only when trained using teacher loss.} Figure \ref{fig:teacher-vs-no} shows that, given a fixed set of examples, the combination of self-play and teacher loss enables models to reach the same level of performance that they would have reached if given unlimited interaction with the oracle on those same examples (i.e. performance of the limited setting is near that of the converged setting). That is, perhaps unsurprisingly, if we can assume that agents are able to ``capture'' a message from the oracle while interacting on an example, there is no additional advantage to continuing to interact with the oracle on that same example. However, as Figure \ref{fig:teacher-vs-no} makes clear, this ability to capture messages is critical in the limited data setting. Without it, no amount of self-play yields performant models, not even in the target role. In the novel role setting, both teacher loss and self-play are essential for learning.

\begin{figure*}[ht!]
     \centering
     \begin{subfigure}[t]{0.25\textwidth}
         \centering
         \includegraphics[width=0.99\textwidth]{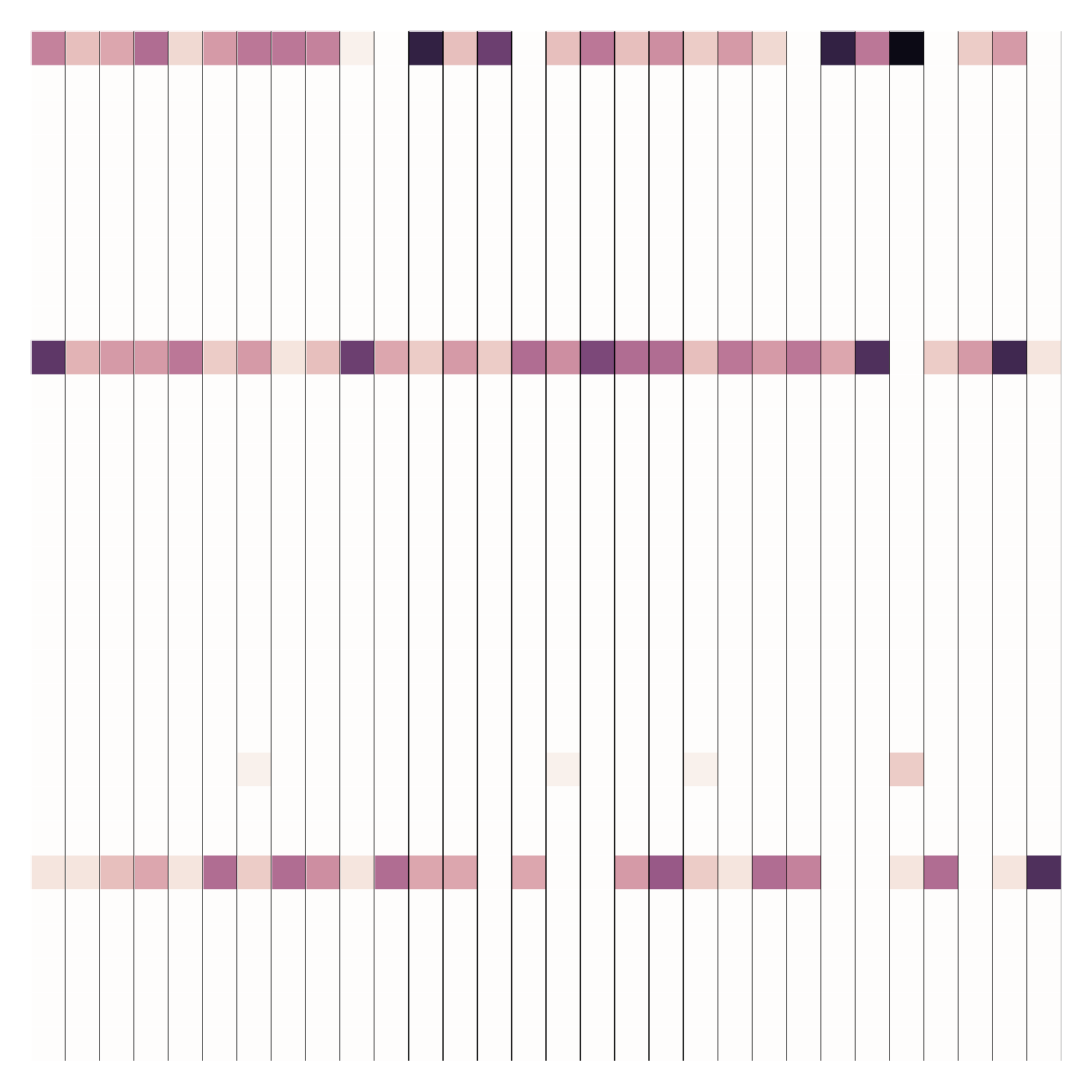}
         \caption{Last super. $M=10$}
     \end{subfigure}%
     ~ 
      \begin{subfigure}[t]{0.25\textwidth}
        \centering
        \includegraphics[width=0.99\textwidth]{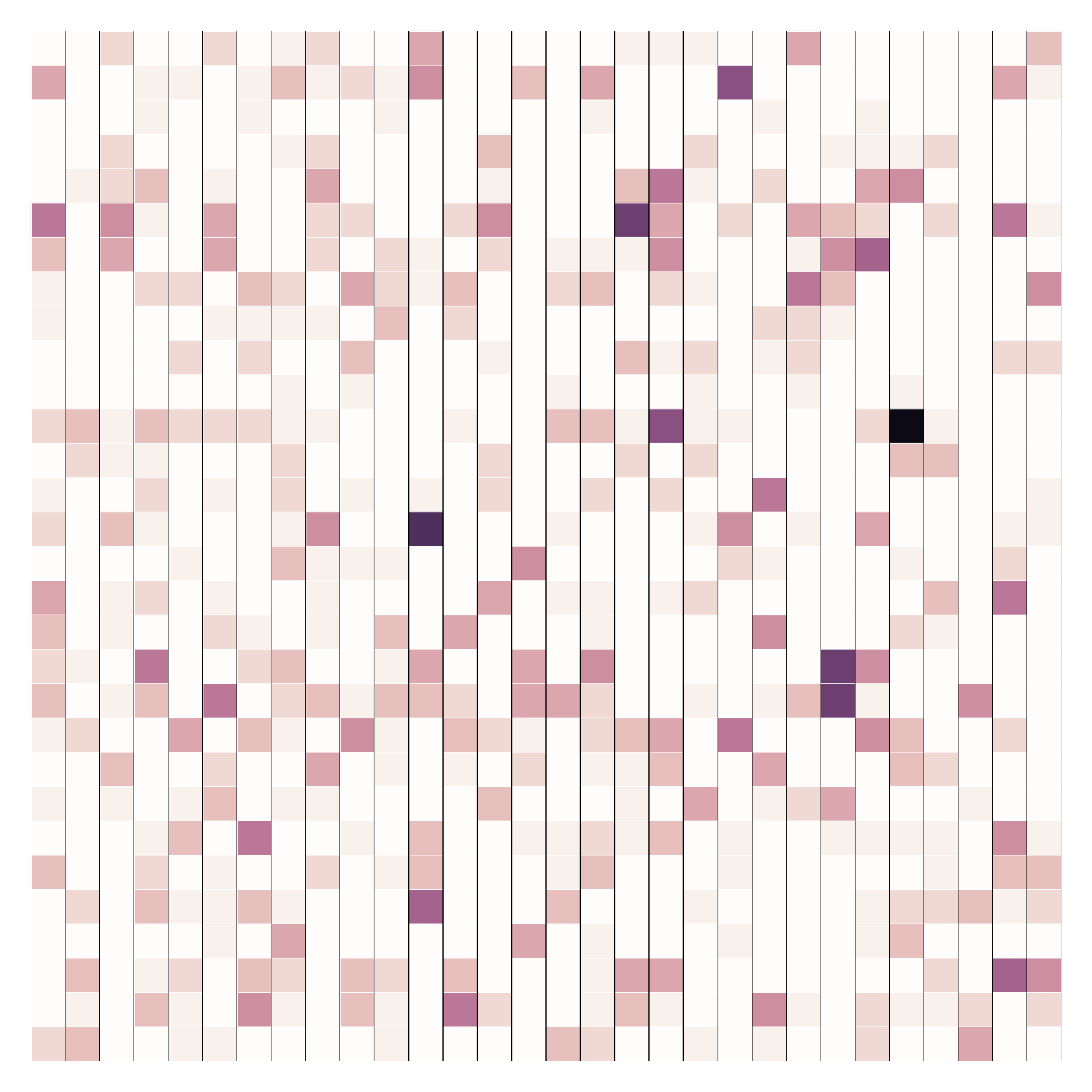}
        \caption{$+$Self-play. $M=10$}
    \end{subfigure}%
    ~
     \begin{subfigure}[t]{0.25\textwidth}
         \centering
         \includegraphics[width=0.99\textwidth]{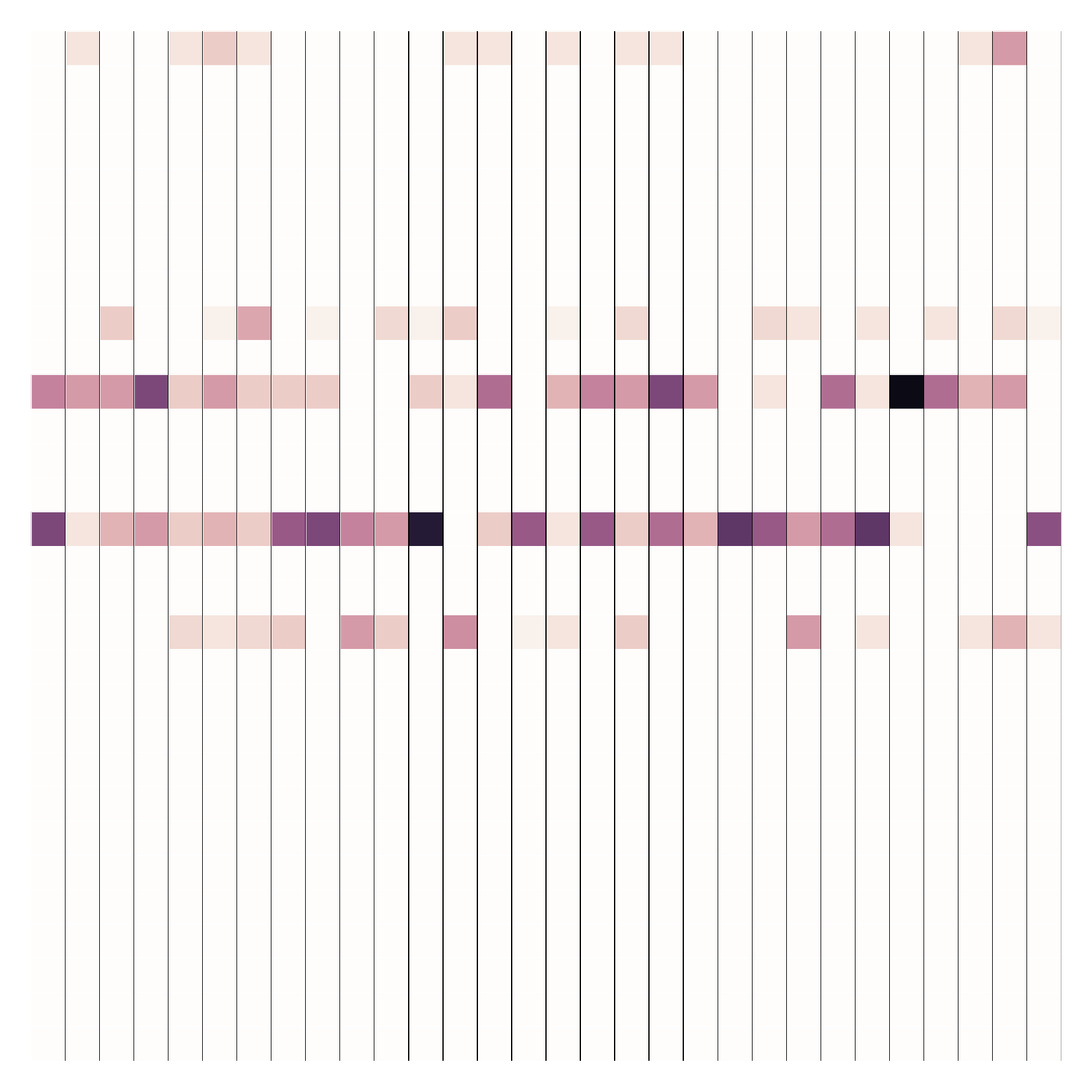}
         \caption{Last super. $M=50$}
     \end{subfigure}%
     ~ 
      \begin{subfigure}[t]{0.25\textwidth}
        \centering
        \includegraphics[width=0.99\textwidth]{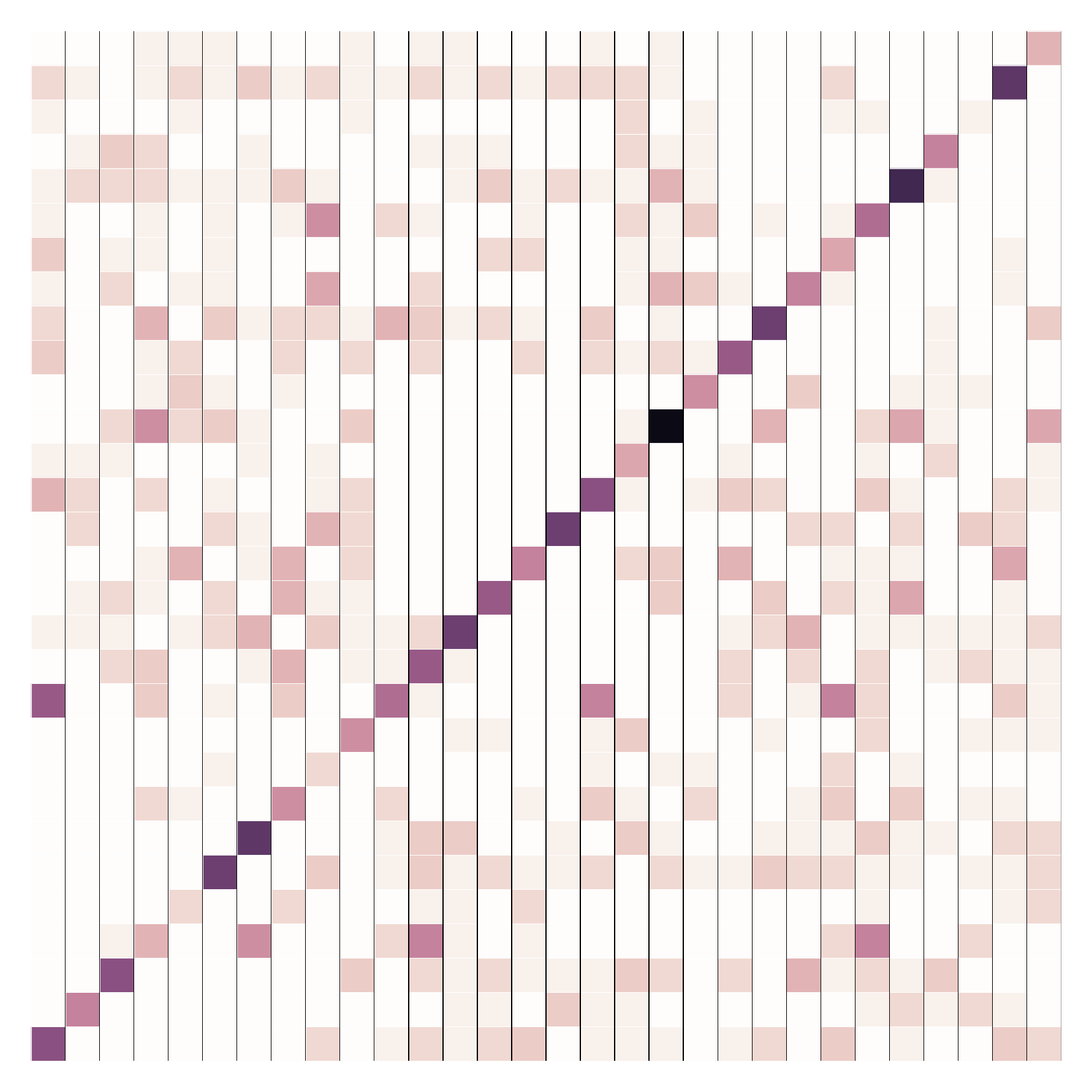}
        \caption{$+$Self-play. $M=50$}
    \end{subfigure}%
    \caption{\textbf{Lexicons learned from limited data.} Visualizations of learned lexicons--i.e. mapping between words ($y$-axis) and attributes ($x$-axis). ``Last super.'' is the state of the lexicon immediately after final interaction with the oracle. ``$+$Self-play'' is after the agent has converged on self-play. Both (a), (c) show that without the additional training time (leveraging both self-play and the teacher loss), the agents over-produce a small set of words. Unsurprisingly, at this level interaction, the agents are far from converged.
    For reference, after only 10 examples, the learning agent reaches 41.8\% accuracy; after 50 examples it reaches 88.3\%. Results are transformer trained as a listener, tested as a speaker, \textit{Shapes} data.}
    \label{fig:oracle-limited-lexicons}
\end{figure*}

\textbf{In the limited data setting, when using self-play and the teacher loss, agents start to perform well with only a handful of examples and quickly develop correct lexicons.}  Table \ref{table:oracle-limited-teacher-accuracies} reports accuracies for transformer agents that use both a teacher and a self-play loss. Even at $M=5$ supervised examples, the agents perform well above the random baseline on the test set. We can see (shown in Figure \ref{fig:oracle-limited-lexicons}) that most of the agent's learning happens during self-play (albeit while using the teacher loss). After 10 examples, the desired lexicon is recognizable; after 50 examples it is sharp.

\begin{table*}[ht!]
    \centering
\begin{tabular}{l|cccc|cccc}
\toprule
 	 & \multicolumn{4}{c|}{\textit{Shapes}} & \multicolumn{4}{c}{\textit{Concepts}} \\
	\cmidrule(lr){2-5} \cmidrule(lr){6-9}
& \multicolumn{2}{c}{Listener} & \multicolumn{2}{c|}{Speaker} & \multicolumn{2}{c}{Listener} & \multicolumn{2}{c}{Speaker} \\
	\cmidrule(lr){2-3} \cmidrule(lr){4-5}\cmidrule(lr){6-7} \cmidrule(lr){8-9}
&    \multicolumn{1}{c}{Target} & \multicolumn{1}{c}{Novel} &    \multicolumn{1}{c}{Target} & \multicolumn{1}{c|}{Novel} &   \multicolumn{1}{c}{Target} & \multicolumn{1}{c}{Novel} &   \multicolumn{1}{c}{Target} & \multicolumn{1}{c}{Novel} \\
\midrule
Random 	& 20.0 & 20.0 & 20.0 & 20.0 & 20.0 & 20.0 & 20.0 & 20.0  \\
M=5       &     33.4 &    30.6 &     34.1 &    45.0 &       32.8 &    22.7 &     26.9 &    33.5 \\
M=10      &     54.6 &    41.8 &     41.7 &    51.1 &       44.5 &    28.2 &     30.5 &    35.9 \\
M=50      &     97.6 &    88.3 &     87.2 &    98.2 &       60.5 &    49.5 &     50.5 &    62.3 \\
M=200     &    100.0 &    99.6 &     99.5 &   100.0 &       78.7 &    81.2 &     68.9 &    77.8 \\
\bottomrule
\end{tabular}
    \caption{\textbf{Limited Supervision Results.} Transformer agent using self-play and teacher loss can both leverage limited direct supervision to perform well in target and in novel roles.}
        \label{table:oracle-limited-teacher-accuracies}
\end{table*}
\section{Discussion \& Conclusion}
We investigate the role of self-play and teacher loss in training agents to communicate from scratch, especially in settings in which access to oracle speakers is limited. In both emergent communication and oracle-training settings, we find strong evidence that agents which self-play can transfer learned information across conversational roles. Lastly, when the self-play loss is coupled with an additional teacher loss--which changes the structure of the reference game--the agents learn to behave in both conversational roles with limited supervised data. Our findings suggest that agents which are directly connected in a graph, linked by agents which communicate, when optimized together, converge towards policies that can interact. This mirrors the results of \cite{graesser2019emergent} which show that communities of learning agents that interact learn to communicate.

\small
\bibliography{main}
\bibliographystyle{main}

\normalsize	
\appendix
\section{Reproducibility and Implementation Details}\label{sec:details}
Our code is available at \url{http://bit.ly/self-play}. 

Each epoch of training, for each item in the dataset, distractors are sampled randomly. Items are represented as bag-of-feature vectors. To improve the stability of results, based on initial pilot experiments, we use reduce learning rate when performance plateaus, as well as using an early stopping criterion to save the best model during training. We measure performance with a rolling average of the downstream accuracies of the last 25 epochs. In our initial pilot experiments, we found that this would better estimate the current performance. Again, we use this rolling average for model selection (best epoch), early stopping (with a patience of 1000 epochs), and for reducing learning rates when performance plateaus. 

We do a hyper-parameter search (\cref{sec:hyperparameters}) with uniform sampling ($N=10$). Test accuracies are averaged across $100$ samples of the test dataset (in each sample the referent remains the same but the distractors are resampled.) Error bars are $95\%$ confidence intervals. In each of our experiments, on average, all dev and test accuracies differ by less than $0.5\%$. The average number of epochs is (1491 emergent, 2026 oracle, 1206 limited) and each epoch takes 0.5-0.8 seconds on Intel Cascade CPUs with 4GB.

The number of emergent jobs run (training and hyperparameter search) is 80.
The number of oracle jobs run (training and hyperparameter search) is 240.
The number of limited jobs run (training with \textit{no} hyperparameter search) is 480.

\subsection{Hyperparameters}\label{sec:hyperparameters}
We perform a hyperparameter search for both the emergent communication and oracle superivision settings, but use fixed  hyperparameters for the oracle supervision limited data setting (to limit our compute). We set it to the setting that work the best in the most settings previously, which we found to be consistent across architectures. See Table \ref{table:hp}. For each hyperparameter setting, we save the best model according to the rolling mean score described in \cref{sec:details}. Then, that model is evaluated across $100$ epochs of development data. The setting with the highest average accuracy (summed over downstream tasks) is selected. 

Although this hyper-parameter search is limited (10 samples), it is important. The Reinforcement Learning is relatively unstable and some settings perform much worse; luckily the better settings are relatively easy to find. The performance difference between HP settings is up to $45\%$, although the average is around ($10\%$ oracle, $25\%$ emergent) with a heavy tail -- s.t. many settings are only a few percentage less than the best setting.

 \begin{table*}[ht!]
   \centering
   \begin{tabular}{@{}ll@{}}
       \toprule 
       \textbf{Search strategy} & uniform sampling (10 samples) \\
       \toprule 
       \textbf{Hyperparameter} & \textbf{Search Space} \\
       \midrule
       \multicolumn{2}{c}{\textit{Optimized Hyperparameters}} \\
       $\beta$ & $0.0$, $\bm{0.001}$ \\
       random seed & 1, $\bm{2}$, 3 \\
        learning rate & $\bm{0.001}$, $0.0005$, $0.0001$ \\
        learning rate scheduler patience & $50, \bm{\infty}$ \\
       \midrule
     \multicolumn{2}{c}{\textit{Fixed Hyperparameters}} \\
     optimizer & RMSProp  \\
     early stopping patience & 1000 \\
     batch size & 64  \\
     number of layers & 1   \\
     hidden dim & 64  \\
     embedding dim & 64  \\
     dropout & 0 \\
     scheduler & ReduceLROnPlateau \\
     \midrule
     \multicolumn{2}{c}{\textit{Transformer Hyperparameters}} \\
     number of attention heads & 1   \\
     transformer feedforward dim & 64 \\
    positional encoding & sinusoidal \cite{vaswani2017attention} \\
    positional encoding dropout & 0.1 \\
      \bottomrule
   \end{tabular}
   \caption{Hyperparameters. Bolded values belong to the setting that performed best across the most oracle settings, and was also used for the subsequent limited data settings. $\beta$ is used to encourage the entropy of agent policies when using the score function estimator. }\label{table:hp}
 \end{table*}

\subsection{Datasets}\label{sec:dataset}
For \textit{Shapes}, we uniformly sample features for all items and then split it into train/test/dev splits randomly. For \textit{Concepts}, we parse the original xml dataset to generate data in a similar form, and then we split it into train/test/dev splits randomly. Both datasets (and the source we used to generate them) are available with our source code.

\section{Rational Speech Acts Algorithm}\label{sec:rsa}
The Rational Speech Acts  \cite{goodman2016pragmatic} algorithm assumes a known literal lexicon that one-to-one maps attributes to words. We use a rational speaker that assumes a literal listener, and selects words to maximize the probability that the listener will select the correct item. We would like the oracle to (1) produce messages of length 10 without exponentially exploding the space of the lexicon, (2) produce messages with a vocabulary $|V| = 100$ smaller than the number of attributes ($597)$, and (3) allow the oracle to use words that correspond to the category of the item. 

Thus, the oracle produces messages with words that either correspond to the raw attributes of the items or the category of the item (like ``Animal'' for a ``Salmon''). The learning agent does not get access to the category information directly in the items, but can learn to co-occuring sets of attributes with the oracle's category words.

To generate messages for all items:
\begin{enumerate}
    \item For each item in the dataset, we add the category of the item to its attributes. The learning agents do not have access to this information.
    \item Use the RSA algorithm to build up a rational speaker with probabilities for each for all the items in the dataset. (The oracle uses both test and train data to inform its messages.)
    \item Produces messages for each item by sampling attributes (argmax) incrementally. Once a attribute has been sampled for an item, we set a mask such that the probability of being sampled is very low ($1 * 10^{-9}$). (Thus, it will be preferred over random attributes, but not before other attributes present in the item.)
    \item Gather counts of the attributes used by the rational speaker across all items.
    \item Select the $|V| = 100$ most common attributes.
    \item Filter out all attributes absent from the $100$ most commmon attributes.
    \item Re-run RSA over the modified attributes.
\end{enumerate}

\paragraph{Speaking} When speaking, the rational speaker oracle produces the message it associates with referent item.

\paragraph{Listening} For each item in context, the rational speaker produces the message it associates with that item. Then, it summarizes each of those messages by summing the words together s.t. it effectively has a bag-of-features representation for each item. It does the same thing for the message it receives from the speaker. Then (just as the learning agents do) it does a dot product with vectors to produce a distribution across the context.

\section{More Limited Oracle Supervision and \textit{Concepts} Results}\label{sec:extra}
\begin{figure*}[t!]
    \centering
    \begin{subfigure}[t]{0.5\textwidth}
        \centering
    \includegraphics[width=\linewidth]{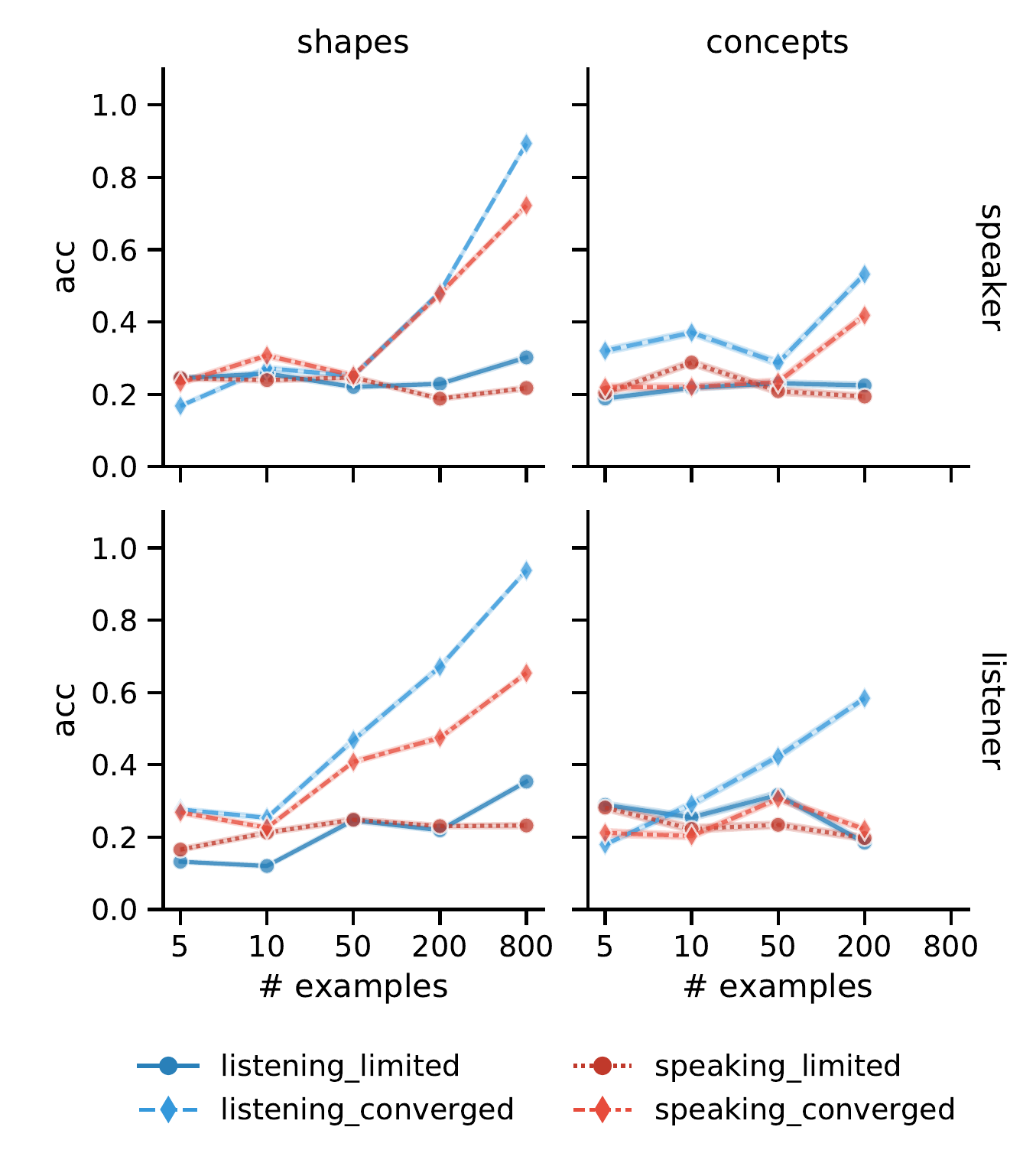}
    \caption{ \textbf{Limited Oracle Supervision; Self-play, No Teacher, Transformer model.} }
    \end{subfigure}%
    ~ 
    \begin{subfigure}[t]{0.5\textwidth}
        \centering
    \includegraphics[width=\linewidth]{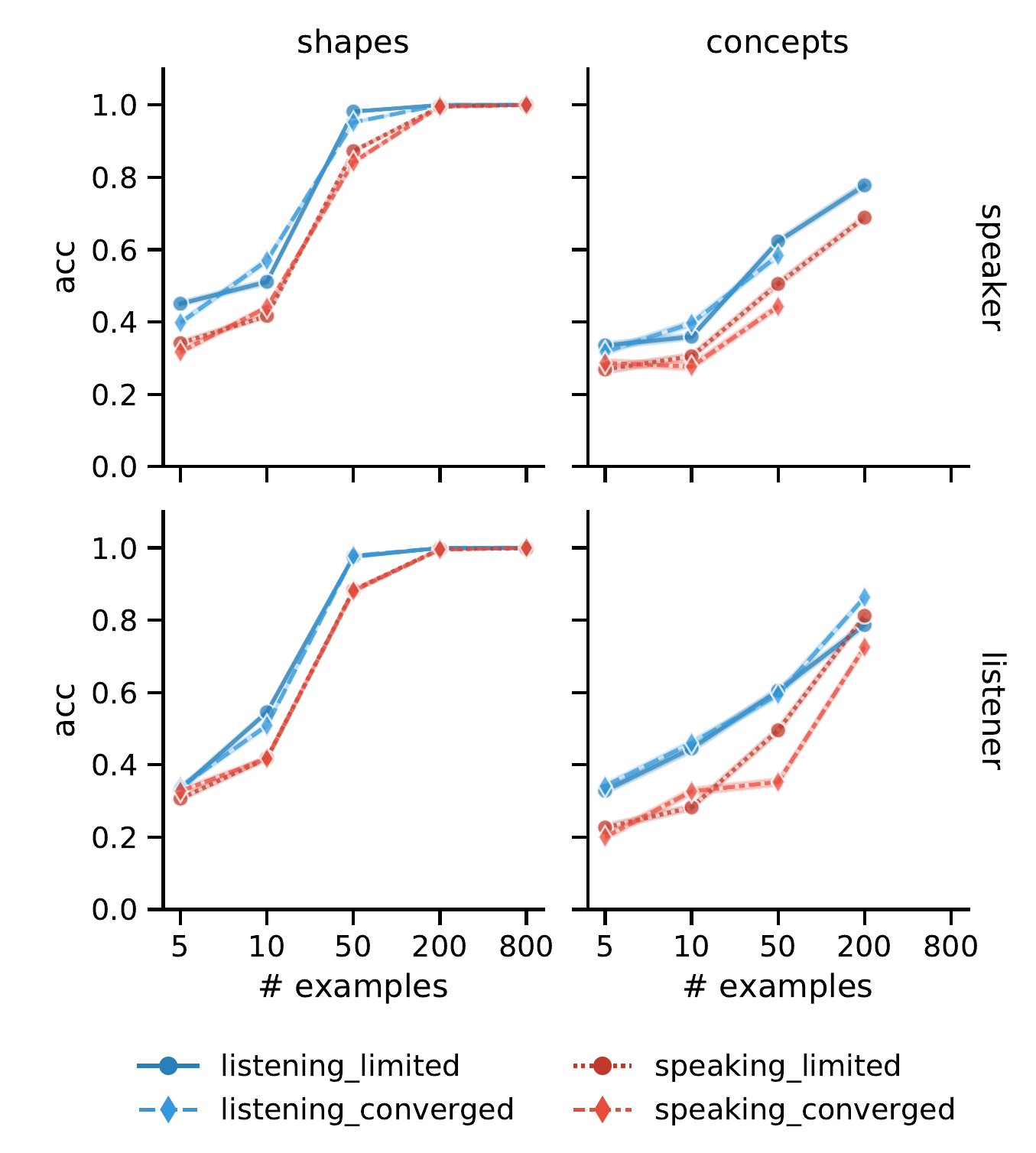}
        \caption{\textbf{Limited Oracle Supervision; Self-play, Teacher, Transformer model.}}
    \end{subfigure}
    ~ \\
    \centering
    \begin{subfigure}[t]{0.5\textwidth}
        \centering
    \includegraphics[width=\linewidth]{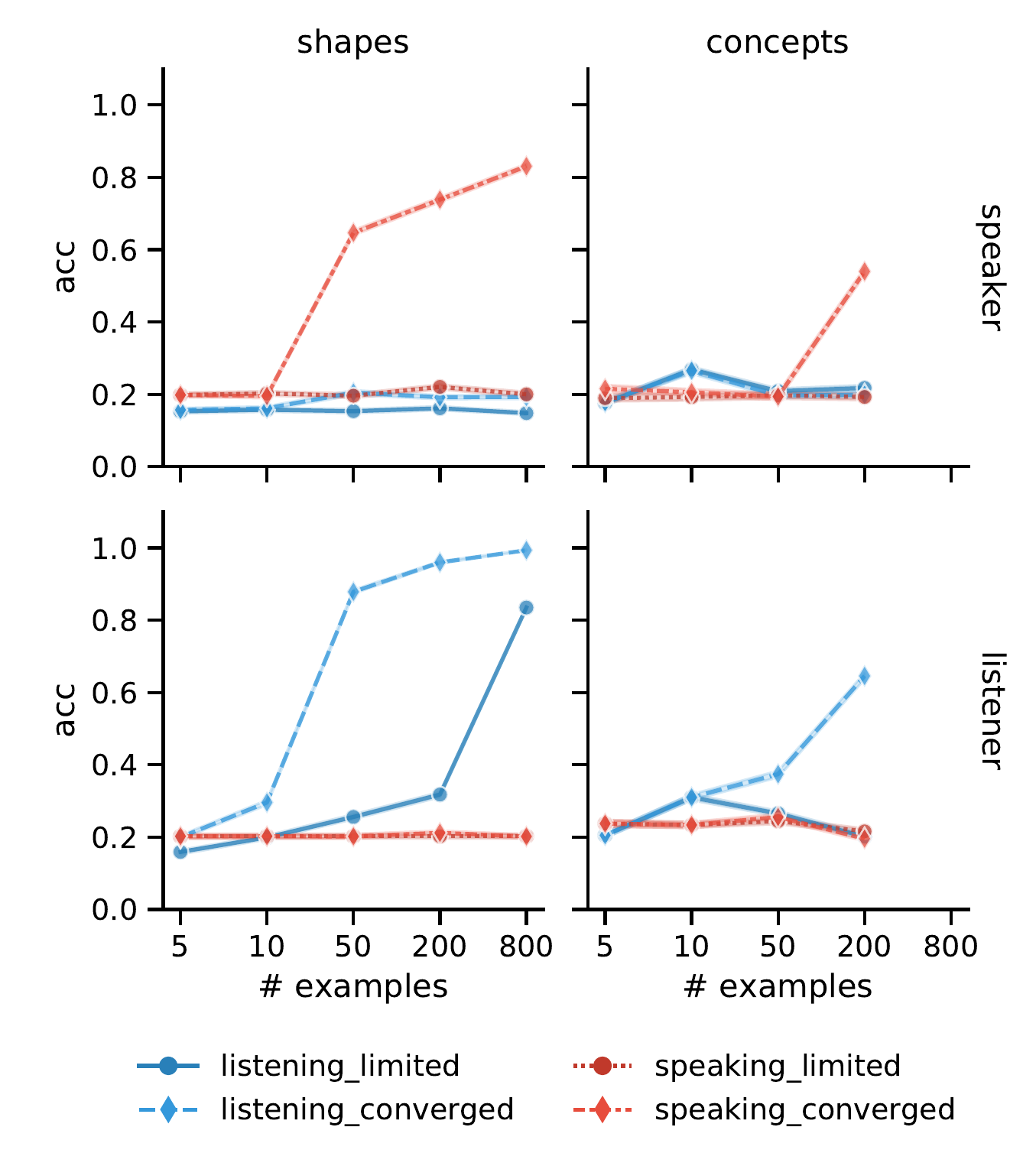}
    \caption{ \textbf{Limited Oracle Supervision; No Self-play, No Teacher, Transformer model.} }
    \end{subfigure}%
    ~ 
    \begin{subfigure}[t]{0.5\textwidth}
        \centering
    \includegraphics[width=\linewidth]{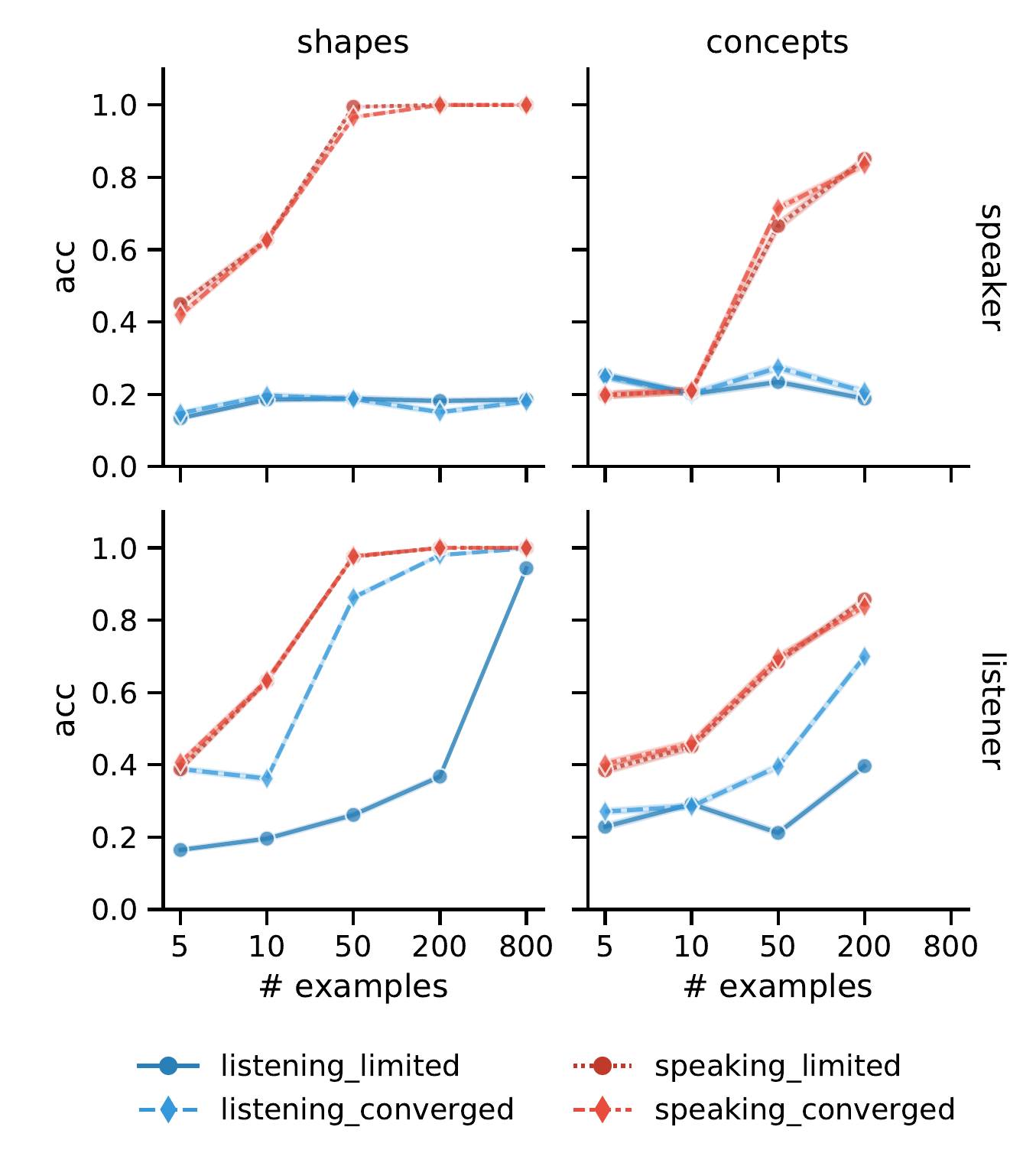}
        \caption{\textbf{Limited Oracle Supervision; No Self-play, Teacher, Transformer model.}}
    \end{subfigure}
    \caption{\textbf{Limited Oracle Supervision, Transformer.}}
    \label{fig:transformer-limited}
\end{figure*}

\begin{figure*}[t!]
    \centering
    \begin{subfigure}[t]{0.5\textwidth}
        \centering
    \includegraphics[width=\linewidth]{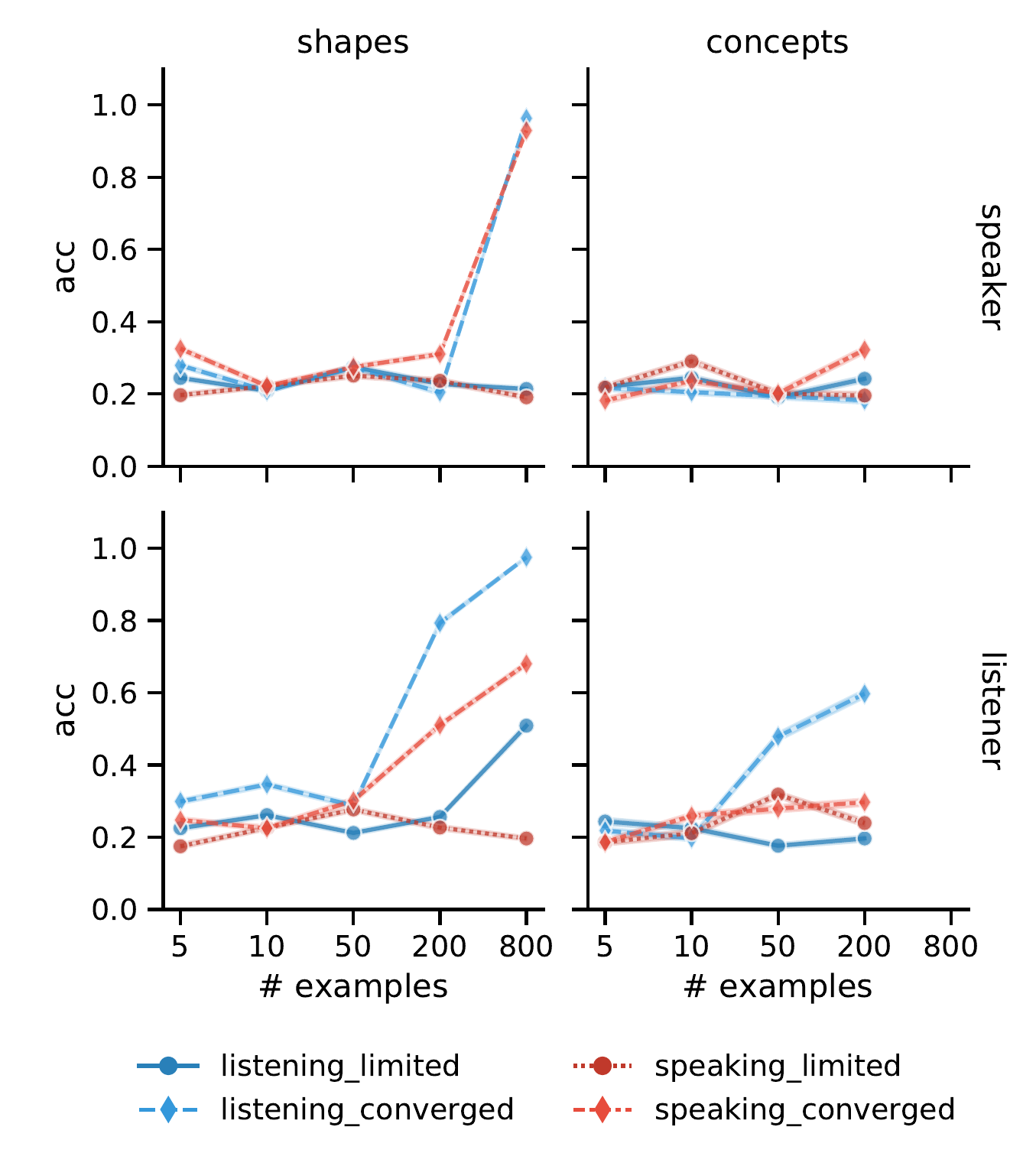}
    \caption{ \textbf{Limited Oracle Supervision; Self-play, No Teacher, LSTM model.} }
    \end{subfigure}%
    ~ 
    \begin{subfigure}[t]{0.5\textwidth}
        \centering
    \includegraphics[width=\linewidth]{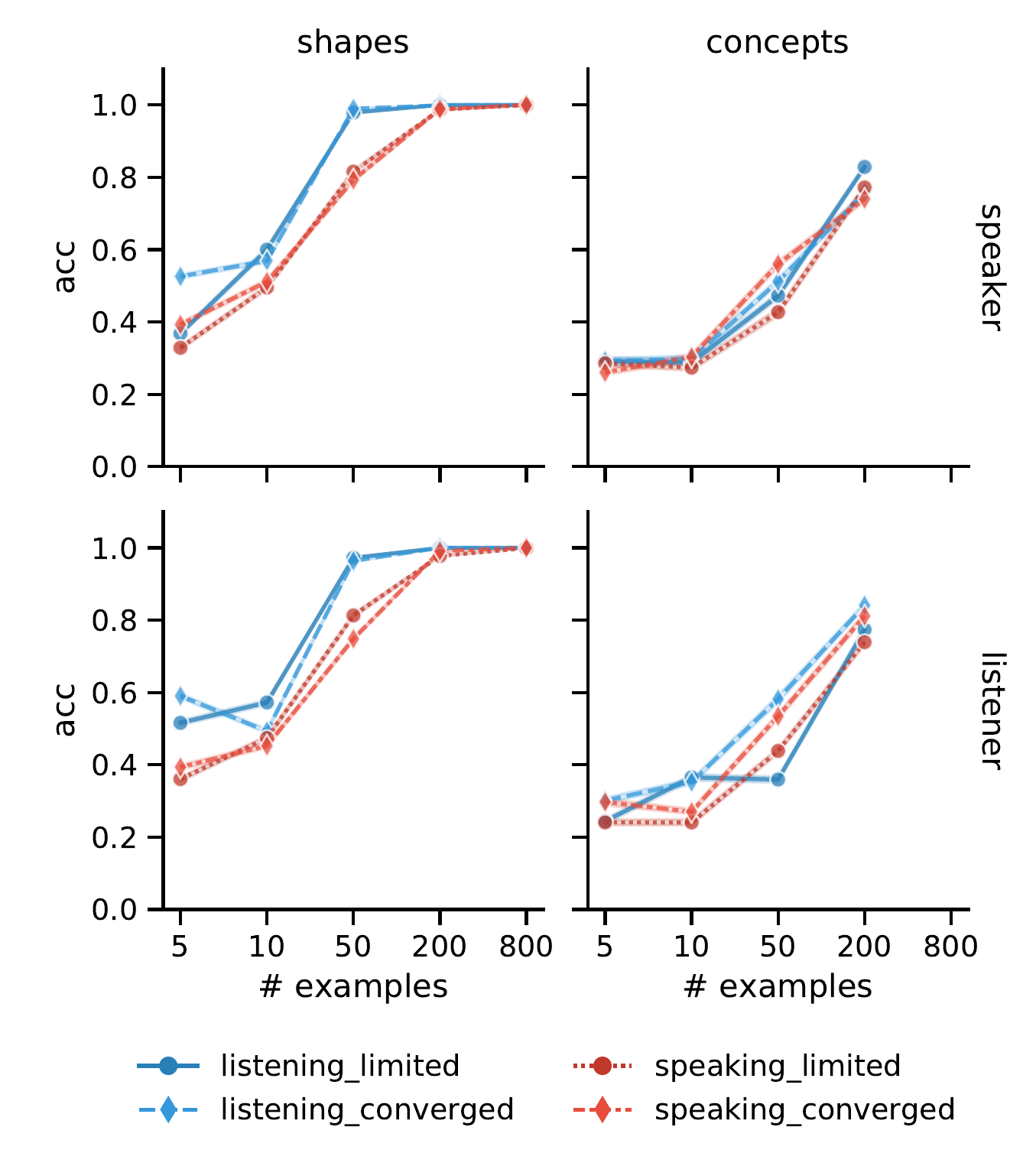}
        \caption{\textbf{Limited Oracle Supervision; Self-play, Teacher, LSTM model.}}
    \end{subfigure}
    ~ \\
        \centering
    \begin{subfigure}[t]{0.5\textwidth}
        \centering
    \includegraphics[width=\linewidth]{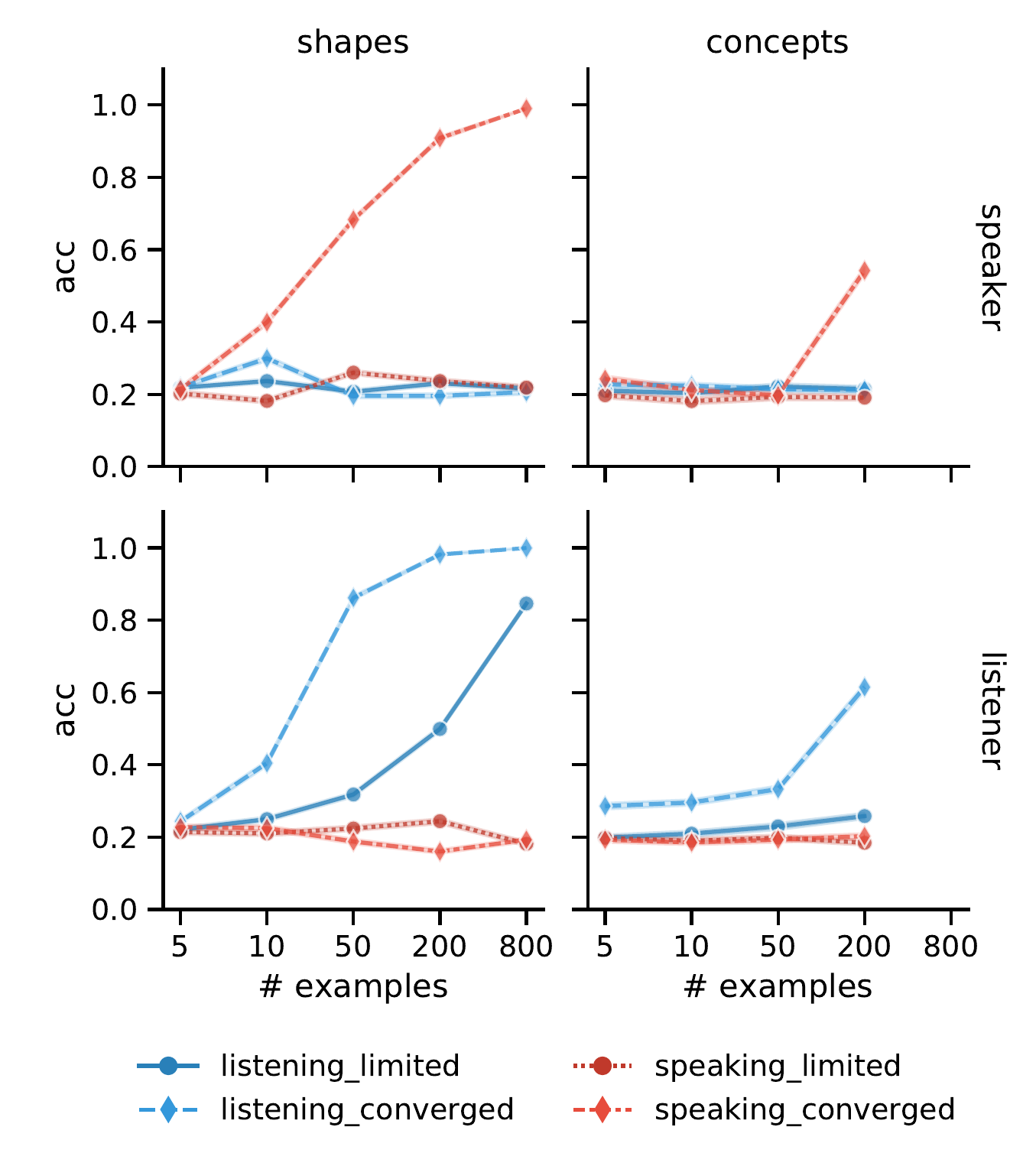}
    \caption{ \textbf{Limited Oracle Supervision; No Self-play, No Teacher, LSTM model.} }
    \end{subfigure}%
    ~ 
    \begin{subfigure}[t]{0.5\textwidth}
        \centering
    \includegraphics[width=\linewidth]{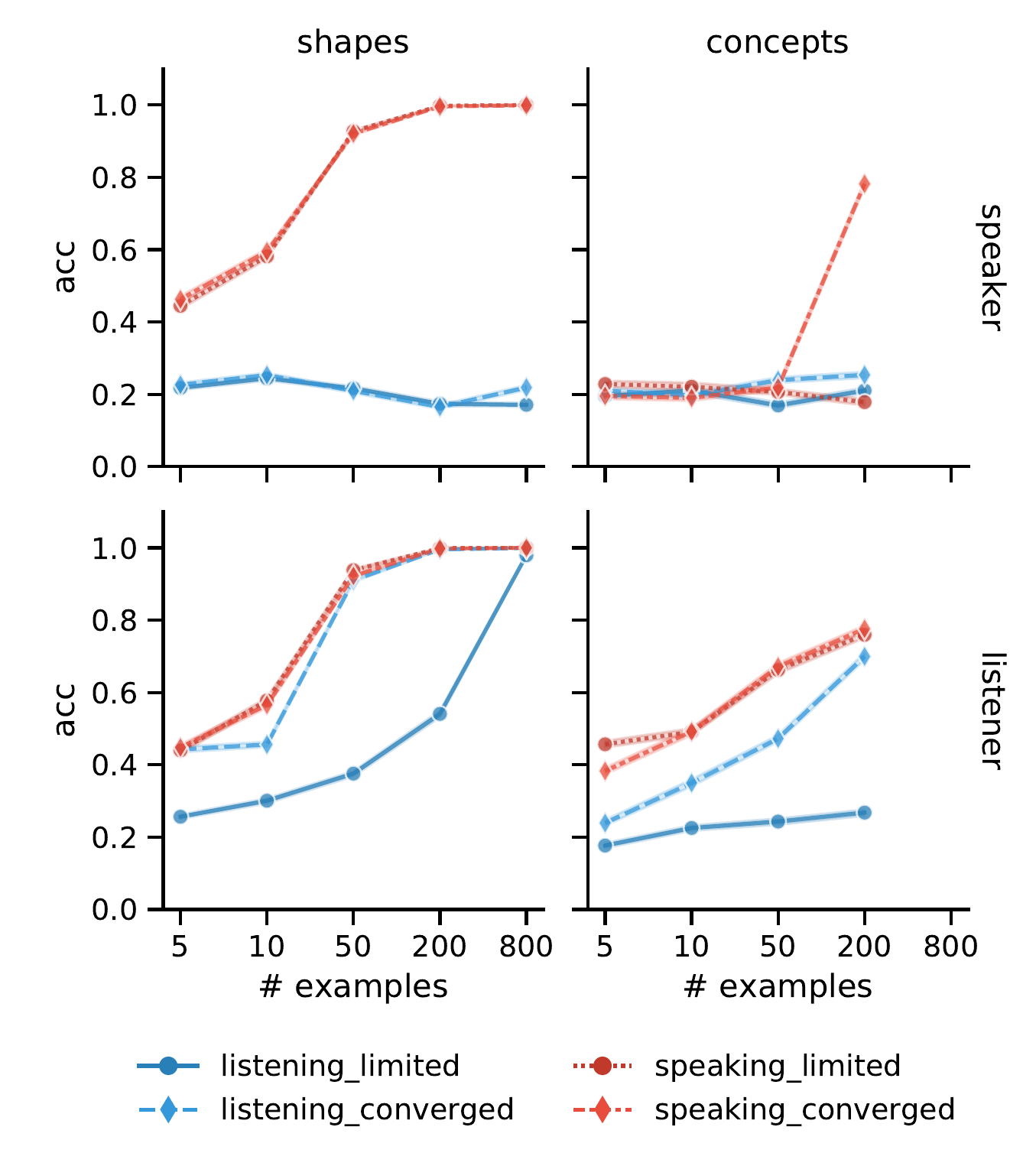}
        \caption{\textbf{Limited Oracle Supervision; No Self-play, Teacher, LSTM model.}}
    \end{subfigure}
    \caption{\textbf{Limited Oracle Supervision, LSTM.}}
    \label{fig:rnn-limited}
\end{figure*}

View the complete results in Table \ref{table:big-limited}. We also present the results visually in Figures \ref{fig:transformer-limited} and \ref{fig:rnn-limited}.

\begin{table*}[ht!]
\tiny
    \centering
\begin{tabular}{llllrrrrrrrr}
\toprule
            &         &          &  & \multicolumn{4}{l}{\textit{Shapes}} & \multicolumn{4}{l}{\textit{Concepts}} \\
            &         &          &  & \multicolumn{2}{l}{Listener} & \multicolumn{2}{l}{Speaker} & \multicolumn{2}{l}{Listener} & \multicolumn{2}{l}{Speaker} \\
            &         &          &  & Target & Novel & Target & Novel &  Target & Novel & Target & Novel \\
model & teacher & self-play & limited &          &         &          &         &            &         &          &         \\
\midrule
rnn & not\_teacher & not\_self-play & 5   &     22.1 &    21.4 &     20.2 &    21.8 &       19.9 &    19.8 &     19.7 &    21.0 \\
            &         &          & 10  &     25.0 &    21.0 &     18.2 &    23.7 &       21.0 &    19.0 &     18.1 &    20.3 \\
            &         &          & 50  &     31.8 &    22.4 &     26.0 &    20.7 &       23.0 &    19.9 &     19.3 &    22.1 \\
            &         &          & 200 &     49.9 &    24.4 &     23.7 &    23.0 &       25.8 &    18.4 &     19.1 &    21.4 \\
            &         & self-play & 5   &     22.5 &    17.5 &     19.7 &    24.4 &       24.3 &    18.6 &     21.8 &    21.9 \\
            &         &          & 10  &     26.1 &    22.6 &     22.2 &    20.8 &       22.5 &    21.1 &     29.0 &    24.5 \\
            &         &          & 50  &     21.2 &    27.7 &     25.1 &    27.4 &       17.6 &    31.8 &     20.1 &    19.0 \\
            &         &          & 200 &     25.6 &    22.7 &     23.6 &    22.8 &       19.7 &    23.9 &     19.5 &    24.2 \\
            & teacher & not\_self-play & 5   &     25.6 &    44.0 &     44.5 &    21.8 &       17.7 &    45.7 &     22.8 &    19.5 \\
            &         &          & 10  &     30.1 &    57.8 &     58.2 &    24.4 &       22.5 &    49.1 &     22.1 &    21.1 \\
            &         &          & 50  &     37.6 &    93.8 &     92.7 &    21.7 &       24.3 &    66.2 &     20.6 &    16.9 \\
            &         &          & 200 &     54.1 &    99.9 &     99.8 &    17.4 &       26.8 &    76.0 &     17.8 &    21.0 \\
            &         & self-play & 5   &     51.6 &    36.1 &     32.9 &    36.8 &       24.4 &    24.1 &     28.6 &    29.2 \\
            &         &          & 10  &     57.2 &    47.4 &     49.5 &    60.0 &       36.5 &    24.1 &     27.4 &    28.8 \\
            &         &          & 50  &     97.2 &    81.3 &     81.6 &    97.9 &       35.9 &    43.8 &     42.7 &    47.3 \\
            &         &          & 200 &    100.0 &    97.8 &     98.7 &   100.0 &       77.4 &    73.9 &     77.2 &    82.9 \\
trans. & not\_teacher & not\_self-play & 5   &     15.9 &    20.2 &     19.8 &    15.3 &       20.5 &    23.7 &     18.9 &    17.6 \\
            &         &          & 10  &     19.9 &    20.2 &     20.2 &    15.7 &       31.0 &    23.4 &     19.2 &    26.8 \\
            &         &          & 50  &     25.6 &    20.2 &     19.6 &    15.3 &       26.5 &    24.4 &     19.7 &    20.9 \\
            &         &          & 200 &     31.8 &    20.2 &     22.1 &    16.1 &       20.1 &    21.6 &     19.3 &    21.7 \\
            &         & self-play & 5   &     13.2 &    16.5 &     24.5 &    24.4 &       28.9 &    28.2 &     20.4 &    18.8 \\
            &         &          & 10  &     12.0 &    21.3 &     23.9 &    25.7 &       25.5 &    22.3 &     28.8 &    21.8 \\
            &         &          & 50  &     24.7 &    24.8 &     24.7 &    22.0 &       31.7 &    23.4 &     20.9 &    23.0 \\
            &         &          & 200 &     21.9 &    23.0 &     18.8 &    22.9 &       18.5 &    19.7 &     19.4 &    22.5 \\
            & teacher & not\_self-play & 5   &     16.5 &    38.8 &     44.9 &    13.4 &       22.9 &    38.5 &     19.8 &    25.3 \\
            &         &          & 10  &     19.6 &    63.2 &     62.7 &    18.5 &       29.1 &    45.0 &     21.0 &    20.1 \\
            &         &          & 50  &     26.2 &    97.5 &     99.5 &    18.8 &       21.2 &    68.5 &     66.6 &    23.4 \\
            &         &          & 200 &     36.8 &   100.0 &    100.0 &    18.2 &       39.7 &    85.7 &     85.1 &    18.8 \\
            &         & self-play & 5   &     33.4 &    30.6 &     34.1 &    45.0 &       32.8 &    22.7 &     26.9 &    33.5 \\
            &         &          & 10  &     54.6 &    41.8 &     41.7 &    51.1 &       44.5 &    28.2 &     30.5 &    35.9 \\
            &         &          & 50  &     97.6 &    88.3 &     87.2 &    98.2 &       60.5 &    49.5 &     50.5 &    62.3 \\
            &         &          & 200 &    100.0 &    99.6 &     99.5 &   100.0 &       78.7 &    81.2 &     68.9 &    77.8 \\
\bottomrule
\end{tabular}
    \caption{Limited Results.}
        \label{table:big-limited}
\end{table*}

\end{document}